\documentclass{article} % For LaTeX2e
\PassOptionsToPackage{numbers, compress}{natbib}
\usepackage{iclr2026_conference,times}

\usepackage{amsmath,amsfonts,bm}

\def\eqref#1{equation~\ref{#1}}
\def\1{\bm{1}}

\DeclareMathAlphabet{\mathsfit}{\encodingdefault}{\sfdefault}{m}{sl}
\SetMathAlphabet{\mathsfit}{bold}{\encodingdefault}{\sfdefault}{bx}{n}

\usepackage{hyperref}
\usepackage{url}
\usepackage{graphicx}
\usepackage{booktabs}
\usepackage{makecell}
\usepackage{amsmath}
\usepackage{amssymb}
\usepackage{mathtools}
\usepackage{amsthm}
\usepackage{wrapfig}
\usepackage[capitalize,noabbrev]{cleveref}

\usepackage{pifont}% http://ctan.org/pkg/pifont
\usepackage{tabu}
\usepackage{multirow}
\usepackage{xcolor,colortbl}
\usepackage{hhline}

\usepackage{enumitem}
\usepackage{tabularx}
\usepackage{nicefrac}       % compact symbols for 1/2, etc.

\usepackage{amsfonts}
\usepackage{color}
\usepackage{algorithm}
\usepackage{algorithmic}

\newtheorem{definition}{Definition}

\usepackage{adjustbox} % added for resizing table

\newcommand{\name}{\textsc{MERGE}}

\title{Massively Multimodal Foundation Models: A Framework for Capturing Interactions with Specialized Mixture-of-Experts}
\author{Xing Han \\
Johns Hopkins University\\
Baltimore, MD 21218, USA \\
\texttt{xhan56@jhu.edu} \\
\And
Hsing-Huan Chung \& Joydeep Ghosh \\
University of Texas at Austin \\
Austin, TX 78712, USA \\
\texttt{\{hhchung,jghosh\}@utexas.edu} \\
\AND
Paul Pu Liang\thanks{Equal Advising.} \\
Massachusetts Institute of Technology \\
Cambridge, MA 02139, USA \\
\texttt{ppliang@mit.edu} \\
\And
Suchi Saria$^*$ \\
Johns Hopkins University\\
Baltimore, MD 21218, USA \\
\texttt{ssaria1@jhu.edu} \\
}

\iclrfinalcopy

\definecolor{ashgrey}{rgb}{0.7, 0.75, 0.71}
\iclrfinalcopy % Uncomment for camera-ready version, but NOT for submission.
\begin{document}

\maketitle

\begin{abstract}
Modern applications increasingly involve many heterogeneous input streams, such as clinical sensors, wearable device data, imaging, and text, each with distinct measurement models, sampling rates, and noise characteristics. We define this as \textit{massively multimodal} setting, where each sensor constitutes a separate modality. As modality counts grow, capturing their complex, time-varying interactions such as delayed physiological cascades between sensors, has becomes essential yet challenging. Mixture-of-Experts (MoE) architectures are naturally suited for this setting since their sparse routing mechanism enables efficient scaling across many modalities. However, existing MoE architectures route tokens based on similarity alone, overlooking the rich temporal dependencies across modalities: this prevents the model from capturing delayed cross-modal effects, leading to suboptimal expert specialization and reduced accuracy.
We propose a framework that explicitly quantifies temporal dependencies between modality pairs across multiple discrete time intervals, defined as delays between an event in one input stream and its manifested effect in another, and uses these to guide MoE routing. A interaction-aware router dispatches tokens to specialized experts based on interaction type. This principled routing enables experts to learn generalizable interaction-processing skills. Experiments across healthcare, activity recognition, and affective computing benchmarks demonstrate substantial performance gains and interpretable routing patterns aligned with domain knowledge.
\end{abstract}

\section{Introduction}
\label{sec:intro}

Multimodal learning has traditionally focused on integrating two or three canonical modalities such as text, image, and audio \citep{baltruvsaitis2018multimodal, liang2024foundations}. Yet real-world applications increasingly involve massively multimodal data: dozens to hundreds of heterogeneous input streams, each with distinct measurement models, sampling rates, noise characteristics, and temporal dynamics \citep{liang2021multibench, liang2022highmmt, soenksen2022integrated}. In healthcare alone, a single patient generates continuous signals from heart rate monitors, pulse oximeters, blood pressure cuffs, ECGs, respiratory sensors, and laboratory analyzers, alongside imaging and clinical notes \citep{johnson2023mimic}. Each stream constitutes a distinct modality requiring different processing strategies \citep{soenksen2022integrated}. As modality counts grow, the space of cross-modal interactions explodes: some pairs provide redundant information, others capture unique signals, and still others exhibit synergistic patterns that emerge only through joint analysis \citep{williams2010nonnegative, liang2023quantifying}. These dependencies often unfold over time with characteristic delays \citep{weissman2012directed, varley2023decomposing}, a predictable time intervals between a cause in one modality and its observable effect in another, further complicating the modeling challenge.
In response, MoE models \citep{shazeer2017outrageously, guo2025deepseek, jiang2024mixtral} offer a natural foundation for this setting. It provides a principled way to allocate computation to modality-specific experts: LIMoE \citep{mustafa2022multimodal} designed multimodal MoE that processes both images and text through shared sparse MoEs with contrastive learning, demonstrating modality specialization through entropy regularization. FuseMoE \citep{han2024fusemoe} addressed the ``FlexiModal'' setting with irregularly sampled and missing modalities, introducing a Laplace gating function with theoretical convergence guarantees superior to softmax routing in multimodal applications. Building on this, Flex-MoE \citep{yun2024flex} proposed a missing modality bank and dual-router design to handle arbitrary modality availability. Hierarchical MoE \citep{nguyen2024expert} demonstrated that Laplace gating at two hierarchical levels eliminates undesirable parameter interactions, accelerating expert convergence in multimodal tasks.

Despite the successful application of MoE in multimodal problems, most existing MoE architectures rely on routers that consider only the similarity between input tokens and experts; such routers only operate on static modalities \citep{riquelme2021scaling, fedus2022switch}, or static features from temporal modalities \citep{xie2025more, zhu2025hierarchical}. However, in real-world applications, using multimodal interaction at a single time point often fails to capture the full landscape of cross-modal relationships, as delayed effects between variables or modalities are common~\citep{zhang2025hierarchical, zhouoracle}. For example, when understanding human communication, a brief eyebrow raise followed 200–400 ms later by rising intonation and a half-smile turns the same words into sarcasm; in medical diagnosis, a slow overnight drift in $\mathrm{SpO_2}$ and respiratory rate, followed hours later by fever and elevated lactate, flags early sepsis. This raises an essential question: \textit{Can we incorporate temporal multimodal interaction to guide MoE training and inference?} To capture such temporal interactions, a successful MoE model should be able to: (1) employ a well-defined quantitative measure of multimodal interaction that accounts for time-delayed interactions between modalities, and (2) incorporate an enhanced MoE architecture capable of leveraging this information during training. With these components, the model can become more proactive and context-aware, leading to improved performance by better understanding complex, time-evolving multimodal processes.

% \todo{can u explain using the 2 examples above what successful temporal moe would look like? and how it would lead to much better performance?} a quantitive interaction measure that captures the whole procedure, the router that leverage this information;  examples...

\begin{figure}
    \centering
    \includegraphics[width=\linewidth]{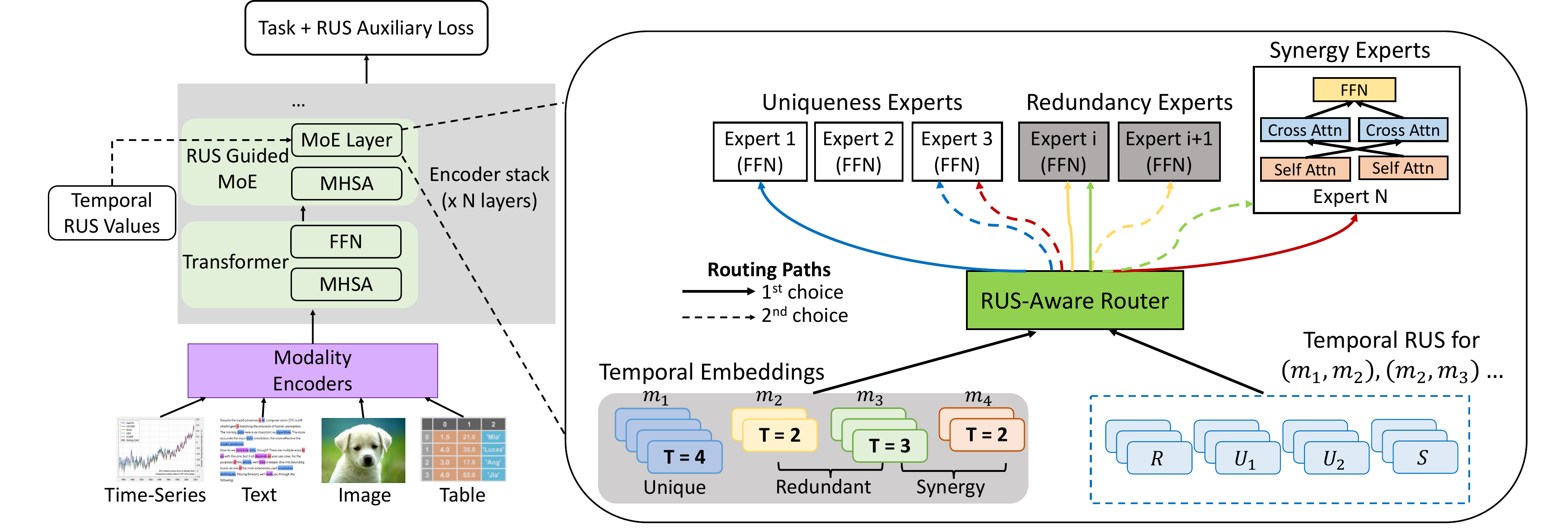}
    \vspace{-2em}
    \caption{Overview of \name. The left panel illustrates the overall architecture, where multimodal inputs are processed through $N$ stacked encoder layers composed of alternating Transformer and MoE blocks. The core innovation of \name~lies in the MoE layers, detailed on the right. The temporal RUS-aware router is the essential part, which leverages temporal multimodal interactions to guide the routing of token embeddings across different time lags. The router determines, based on interaction dynamics, which modality pairs should (or should not) be routed to the same expert, thereby enabling more principled and interpretable expert specialization. As an example, $m_2$ (yellow) and $m_3$ (green) exhibit high redundancy according to their temporal RUS values; therefore, the router is more likely to assign them to the same expert (yellow and green arrows).}
    \label{fig:overview}
\end{figure}

Motivated by this challenge, we propose \textbf{M}assively-multimodal \textbf{E}xpert \textbf{R}outing for \textbf{G}eneralized \textbf{E}xchange (\name), a novel framework that leverages temporal multimodal interactions to guide the routing process of MoE. The overall framework is illustrated in Figure \ref{fig:overview}. We begin by introducing a method to explicitly compute temporal multimodal interactions (\textbf{R}edundancy, \textbf{U}niqueness, \textbf{S}ynergy, or RUS) between input modalities over time with respect to the target outcome. To handle high-dimensional and temporal data, we design a scalable approach based on multi-scale BATCH estimator to compute multi-step temporal interactions
efficiently. The resulting estimates are then used to guide the MoE training process. Specifically, we design an interaction-aware router that incorporates the context of temporal RUS sequences and dynamically routes tokens to experts based on these temporal interactions. This routing mechanism is further reinforced using auxiliary loss functions during training. The core of MERGE consists of two components: (1) computing temporal RUS (section \ref{sec:trus}), and (2) using the resulting RUS sequences to guide MoE training (section \ref{sec:moe}). We demonstrate the effectiveness of \name\ in two ways: first, by analyzing the insights provided by temporal RUS and showing that they capture meaningful and application-relevant patterns; and second, by illustrating that the learned routing patterns align with our expectations and lead to significant performance improvements across a diverse set of multimodal tasks.

% \vspace{-0.5em}
\section{Related Works}
\label{sec:related}
\vspace{-0.8em}
\textbf{Multimodal Interactions} define the degrees of commonality between modalities and the ways they combine to provide new information for a task~\citep{liang2024foundations}. A core problem lies in understanding the nature of how modalities interact and modeling these interactions using data-driven methods. The study of multimodal interactions has involved semantic definitions based on research in multimedia~\citep{marsh2003taxonomy}, verbal and nonverbal communication~\citep{partan2005issues,flom2007development,ruiz2006examining}, social interactions~\citep{mai2019divide, jung2018multi}, and instruction tuning \citep{shan2025mint}. These have also inspired statistical methods to quantify multimodal interactions from unimodal predictions~\citep{mazzetto21a}, trained model weights and activations~\citep{sorokina2008detecting,tsang2018detecting,tsang2019feature,hessel2020emap}, feature selection~\citep{ittner2021feature,yu2003efficiently,yu2004efficient,auffarth2010comparison}, and information theory~\citep{williams2010nonnegative,bertschinger2014quantifying}. For information theory-based methods, recent studies have investigated extensions to continuous \citep{pakman2021estimating, ehrlich2024partial} or Gaussian \citep{venkatesh2023gaussian} distributions, many modality decomposition \citep{varley2024generalized}, sample-level quantifications \citep{yang2025efficient}, and enable accurate estimation from large-scale multimodal datasets \citep{liang2023quantifying,liang2024multimodal}. Recently, \citet{varley2023decomposing} introduced a framework for modeling multimodal interactions over time. While insightful, the approach has limitations: it does not naturally extend to continuous variables, and its lattice-based structure does not scale beyond small systems, limiting its practicality for large-scale multimodal applications.

% \todo{this is copied from past work pls modify and add latest work esp in information theory estimation, extension to many modalities, high-dimensional data, temporal modalities, fine-grained data etc.}

% \textbf{Partial Information Decomposition (PID)} \citep{williams2010nonnegative, bertschinger2014quantifying} has emerged as a formal way to quantify multimodal interactions by measuring how the total information between two modalities $(x_1,x_2)$ useful for a task $y$ can be decomposed into redundant ($R$), unique ($U$), and synergistic ($S$) parts. Redundancy measures the common information between two modalities, uniqueness measures the useful information in a modality not present in the others, and synergy measures the new information that arises only when both modalities are fused. 

%employ a binary label agreement method based on unimodal classifiers to approximate multimodal interactions. This approach is a simplistic surrogate for PID, which quantifies interactions at the bit level using mutual information. However, such approximations are inherently imprecise and highly dependent on the performance of the unimodal classifiers. If a unimodal classifier fails, it can produce spurious agreement, leading to incorrect categorization of the interaction type.

\textbf{PID and Multimodal Interaction MoEs.} Partial Information Decomposition (PID) \citep{williams2010nonnegative, bertschinger2014quantifying} provides a principled framework for quantifying multimodal interactions by decomposing the total information into redundant ($R$), unique ($U$), and synergistic ($S$) components. Redundancy captures the shared information across modalities, uniqueness measures modality-specific contributions, and synergy reflects information that emerges only when modalities are combined. To the best of our knowledge, PID has not yet been exploited to improve multimodal MoEs. In contrast, existing multimodal MoE approaches typically rely on heuristic expert assignments \citep{shazeer2017outrageously, he2021fastmoe, han2022dynamic, akbarian2024quadratic}, which often result in opaque and difficult-to-interpret routing decisions \citep{liu2024deepseek,shen2024mome}. Multimodal interaction provides a natural and principled way to enhance expert specialization. For example, \citet{yu2023mmoe} introduced the Mixture of Multimodal Interaction Experts, where specific experts are assigned to process predetermined types of modality interactions. Recently, \citet{xin2025i2moe} extended this idea by incorporating interaction-type categorization into the MoE training process and expanding the framework to support more than two modalities.
However, these designs make the number of experts tightly coupled to the number of modalities, which limits scalability. A more natural and flexible solution is to use multimodal interactions to guide expert routing, enabling the model to dynamically determine which modalities should (or should not) be processed together. Moreover, both works rely on a binary label agreement method based on unimodal classifiers to approximate the PID-type multimodal interactions, making the approach heavily dependent on the performance of the individual classifiers. Additionally, this method captures only static interactions, overlooking the temporal and continuous nature of real-world multimodal data.

\section{RUS-Guided MoE for Massively Multimodal Learning} 

We introduce \name, our framework for guiding MoE with temporal multimodal interactions in massively multimodal settings. The approach consists of two parts: (1) capturing temporal multimodal interactions, and (2) leveraging these quantified interactions to inform the training of MoE.

% \paul{pls be consistent with terminology everywhere - dynamic vs temporal vs time-varying vs time-delayed, dependencies vs interactions are all used}

\subsection{Capturing Temporal Multimodal Interactions}
\label{sec:trus}

We first present our methodology for capturing temporal multimodal interactions. We detail the procedure for computing temporal RUS across different settings, describe how to efficiently estimate these interactions, and provide potential insights that can be derived from the temporal RUS patterns.

\textbf{Formulation of temporal RUS.}
Capturing information interactions over time remains an important yet underexplored problem. \textit{Our goal is to characterize how past and present values interact across time for a task.} For example, in the medical domain, one may wish to understand how the interactions of past treatments influence a patient’s health outcomes; in activity recognition, it is crucial to capture how past motions of individual body parts (e.g., arms and legs) contribute to predicting future overall motion. However, the existing PID framework (Eq. \ref{eqn:ri}–\ref{eqn:si} in Appendix \ref{sec:trus_supp}) is computed using standard mutual information, which only captures static interactions and cannot be directly extended to the temporal setting. To this end, we build on directed information \citep{weissman2012directed}, which enables scalable modeling of temporal interactions while maintaining alignment with PID. 
Directed information respects temporal relationships by considering information flow from past to present, which enables PID analysis across multiple time lags, providing insights into both short-term and long-term influences. We begin by defining multi-source directed information, which quantifies the flow of information from multiple input modalities to the target variable over time.

% sequence x1 1...T x2 1...T, target Y. our goal is to define multimodal interaction between 2 sequences and Y. this is difficult because....
% we use directed info... transfer entropy.. just present multiple source
% now we apply pid. explain pid. explain how to adapt pid. naive way of applying pid. doesnt work.
% key insight! with some assumptions eg. time lag contribution... now we can apply pid efficiently. (key contribution)
% formal equations of pid with time lag

\begin{wrapfigure}[12]{r}{0.42\textwidth}
\vspace{-2.1em}
\begin{center}
    \includegraphics[width=.43\textwidth]{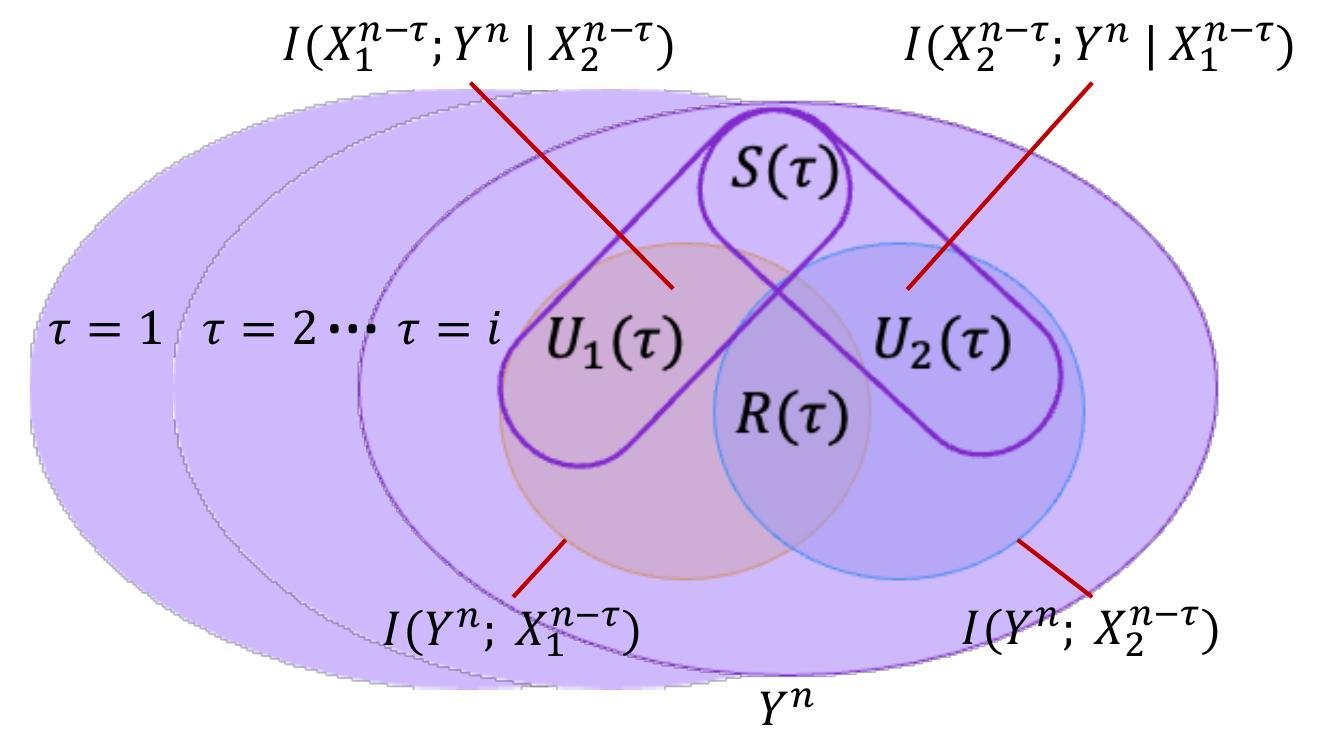}
\end{center}
\vspace{-1.1em}
\caption{Decomposed directed information components across time lag $\tau$.}
\label{fig:venn}
\end{wrapfigure}

\begin{definition}[Multi-Source Directed Information]
Given multiple source processes $X_1^{i-1} = (X_{1,1}, X_{1,2}, \ldots, X_{1,i-1}), ~X_2^{i-1} = (X_{2,1}, X_{2,2}, \ldots, X_{2,i-1})$ and the target process $Y^{i-1} = (Y_1, Y_2, \ldots, Y_{i-1})$, the directed information from $(X_1, X_2)$ to $Y$ over $n$ steps with time lag $\tau$ is:
\begin{equation}
\mathrm{DI}(\tau) = I(X_1^{n-\tau}, X_2^{n-\tau} \to Y^n) = \sum_{t=\tau+1}^{n} I(Y_t; X_{1,t-\tau}, X_{2,t-\tau} | Y^{t-1}).
\label{eq:di}
\end{equation}
\end{definition}
% \begin{definition}[Transfer Entropy]
% The transfer entropy from source $X$ to target $Y$ with lag $\tau$ at time $t$ is:
% \begin{equation}
% \mathrm{TE}_{X \to Y}(t, \tau) = I(Y_t; X_{t-\tau} | Y_{t-1})
% \label{eq:te}
% \end{equation}
% This measures the information flow from $X$'s past to $Y$'s present at time $t$, conditioning on $Y$'s own past.
% \end{definition}
As shown in Figure \ref{fig:venn}, at each time lag $\tau$, $\mathrm{DI}(\tau)$ can be decomposed into four components based on the conditional joint distribution $P_{\tau}(x_1, x_2, y) = P(X_{1}^{n-\tau} = x_1, X_{2}^{n-\tau} = x_2, Y^n = y | Y^{n-1})$, where $\mathrm{DI}(\tau) = R(\tau) + U_{1}(\tau) + U_{2}(\tau) + S(\tau)$: each of its component quantifies interaction at $\tau$:
\begin{align}
R(\tau) &= \min_{Q_{\tau} \in \Delta_{\tau}} I_{Q_{\tau}}(Y^n; X_{1}^{n-\tau}) + I_{Q_{\tau}}(Y^n; X_{2}^{n-\tau}) - I_{Q_{\tau}}(Y^n; X_{1}^{n-\tau}, X_{2}^{n-\tau}), \label{eqn:inter_ri} \\
U_{1}(\tau) &= I_{Q_{\tau}^*}(Y^n; X_{1}^{n-\tau}) - R(\tau), \quad U_{2}(\tau) = I_{Q_{\tau}^*}(Y^n; X_{2}^{n-\tau}) - R(\tau), \label{eqn:inter_ui} \\
S(\tau) &= I_{P_{\tau}}(Y^n; X_{1}^{n-\tau}, X_{2}^{n-\tau}) - I_{Q_{\tau}^*}(Y^n; X_{1}^{n-\tau}, X_{2}^{n-\tau}), \label{eqn:inter_si}
\end{align}

% \paul{define R, U, S with respect to tau, and do we refer to figure 2 from main text?}

where $\Delta$ represents the probability simplex and $\Delta_{\tau}$ is the set of marginal-matching distributions defined as $\Delta_{\tau} := \{ Q_{\tau} \in \Delta : Q_{\tau}(x_i, y) = P_{\tau}(x_i, y), ~ \forall y \in Y, x_i \in X_i, i \in \{1,2\} \}$; it characterizes the set of joint distributions $Q_{\tau}(x_1, x_2, y)$ that preserve the time-specific bivariate marginals $P_{\tau}(x_1, y)$ and $P_{\tau}(x_2, y)$ while allowing the coupling between $X_{1}$ and $X_{2}$ to vary; $Q_{\tau}^* = \arg\min_{Q_{\tau} \in \Delta_{\tau}} I_{Q_{\tau}}(X_{1}^{n-\tau}; X_{2}^{n-\tau} | Y^n)$ is the optimal distribution that has \textit{eliminated the synergistic information} between $X_1$ and $X_2$. After further simplifying the formulation, we obtain
\begin{align}
R(\tau) &= \max_{Q_{\tau} \in \Delta_{\tau}} I_{Q_{\tau}}(X_{1}^{n-\tau}; X_{2}^{n-\tau}; Y^n), \label{eqn:temp_ri} \\
U_{1}(\tau) &= \min_{Q_{\tau} \in \Delta_{\tau}} I_{Q_{\tau}}(X_{1}^{n-\tau}; Y^n | X_{2}^{n-\tau}), \quad U_{2}(\tau) = \min_{Q_{\tau} \in \Delta_{\tau}} I_{Q_{\tau}}(X_{2}^{n-\tau}; Y^n | X_{1}^{n-\tau}), \label{eqn:temp_ui} \\
S(\tau) &= I_{P_{\tau}}(X_{1}^{n-\tau}, X_{2}^{n-\tau}; Y^n) - \min_{Q_{\tau} \in \Delta_{\tau}} I_{Q_{\tau}}(X_{1}^{n-\tau}, X_{2}^{n-\tau}; Y^n). \label{eqn:temp_si}
\end{align}
Note that, $X_1$ and $X_2$ do not necessarily have the same time lag $\tau$, and exhaustively enumerating all pairwise combinations of $\tau$ across modalities would further increase memory costs during model training. To ensure efficiency, we design $X_1$ and $X_2$ to be aligned at the same time lag, while noting that the framework naturally extends to cross-lagged interactions across different time steps.

\begin{figure}
    \centering
    \includegraphics[width=0.85\linewidth]{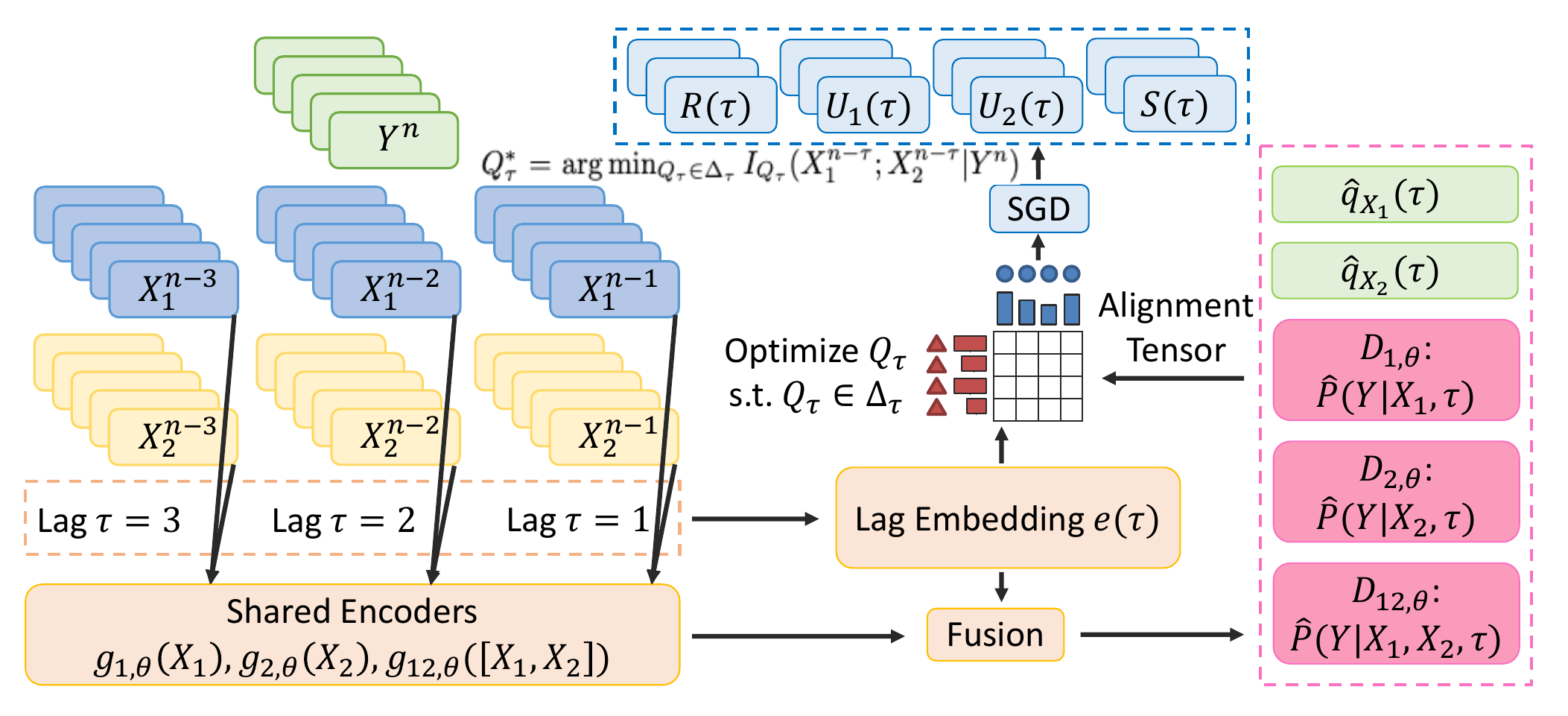}
    % \vspace{-1em}
    \caption{Schematic overview of the multi-scale BATCH estimator. The procedure consists of four stages: (1) encoding empirical datasets $\mathcal{D}_\tau = \big\{(X_{1}^{n-\tau}, X_{2}^{n-\tau}, Y^n)\big\}_{\tau=1}^3$ with shared encoders $g$, with each lag $\tau$ further embedded by an encoder $e$ to produce $e(\tau)$, (2) training lag-conditioned discriminators $D_{1,\theta}, D_{2,\theta}, D_{12,\theta}$ to estimate $\hat{P}$, together with MLPs to generate embeddings for $\hat{q}$ at each $\tau$, (3) updating the alignment tensor to enforce marginal distribution matching between $Q_{\tau}$ and $\hat{P}$, yielding the optimized distribution $Q_{\tau}^*$, and (4) decomposing the resulting estimates of $Q_{\tau}^*$ and $\hat{P}$ into redundancy, uniqueness, and synergy sequence across all time lags.}
    % \todo{in the figure, not clear the Q is for what time steps? and the RUS is for the whole sequence or which time steps? and what are the D1, D2, D12? what does rescale and align mean? what is lag embedding? try to walk the reader through all parts of the figure using the caption }  }
    \label{fig:batch_estimator}
\end{figure}

\textbf{Efficient computation of temporal RUS in high dimensions.}
% We will discuss how RUS at each time stamp can be estimated simultaneously.
Computation of PID was limited to discrete and small support \citep{bertschinger2014quantifying, griffith2014quantifying}, or continuous but low-dimensional variables \citep{pakman2021estimating}. The BATCH estimator \citep{liang2023quantifying} offers an effective solution for handling high-dimensional distributions: it parameterizes the distribution components of interest using neural networks and approximates the true distribution based on subsampled batches.
% \todo{this whole paragraph needs to be much more formal.. give equations for the target terms..what are the models and parameters.. how to learn and what objective..}
The direct way to obtain temporal RUS values is to compute Eq. \ref{eqn:temp_ri}–\ref{eqn:temp_si} for each $t \in [n-\tau, n]$, where all the distributional components of $P_{\tau}$ and $Q_{\tau}$ are estimated using the BATCH estimator. However, this can introduce significant computational overhead, as the optimization required to obtain $Q_{\tau}^*$ may need to be repeated at every step. 
To leverage the multitask nature of neural network backbones, we enhance the BATCH estimator by formalizing a multi-scale approach that trains a single model to predict the temporal RUS at multiple time lags, as shown in Figure \ref{fig:batch_estimator}.

For each lag $\tau$, we construct the empirical dataset $\mathcal{D}_\tau = \big\{(X_{1}^{n-\tau}, X_{2}^{n-\tau}, Y^n)\big\}$, with $\tau$ chosen such that $n - \tau > 1$. We first estimate distributions $P_{\tau}(Y|X_1), P_{\tau}(Y|X_2), P_{\tau}(Y|X_1, X_2)$ by training lag-conditioned discriminators $D_1$, $D_2$, and $D_{12}$ for all $\tau$. For example, $D_{12}$ is defined as
\begin{equation}
    \hat{P}(Y|X_1,X_2, \tau) = D_{12,\theta}\big(\phi(g_{12,\theta}([x_1; x_2]), e(\tau))\big),
    \label{eqn:discriminator}
\end{equation}
where $g_{12,\theta}(\cdot)$ is the encoder, $\phi$ is the fusion operator, and $e(\tau)$ is learnable lag embeddings that encode temporal relationships. We then construct the optimal distribution $Q_{\tau}^*$ that satisfies $Q_{\tau}^* = \arg\min_{Q_{\tau} \in \Delta_{\tau}} I_{Q_{\tau}}(X_{1}^{n-\tau}; X_{2}^{n-\tau} | Y^n)$. This is achieved by constructing the alignment tensor $\text{align}_{\tau} \in \mathbb{R}^{N \times N \times C}$ that measures the compatibility between samples $i$ and $j$ from modalities $X_1$ and $X_2$, within class $k$, at lag $\tau$. Here, $N$ represents the batch size and $C$ the number of classes.
\begin{equation}
\text{align}_{\tau}[i,j,k] = \exp\left(\frac{\hat{q}_{X_1}^{(i,k,\tau)} \cdot \hat{q}_{X_2}^{(j,k,\tau)}}{\sqrt{d}}\right), ~\text{where}~ \hat{q}_{X_{\cdot}}^{(i,k,\tau)} = \text{NN}_{\cdot}(\phi(g_{\cdot,\theta}(x_{\cdot,i}), e(\tau)))_k.
\label{eqn:align}
\end{equation}
The alignment tensors are subsequently normalized via the Sinkhorn–Knopp algorithm \citep{knight2008sinkhorn} to enforce marginal-matching constraints, yielding the optimal distribution $Q_{\tau}^*$. We perform this step for all $\tau$ simultaneously by leveraging the efficient parallelism of tensor operations. Using Eq. \ref{eqn:inter_ri}–\ref{eqn:inter_si}, we can then compute temporal RUS values directly on high-dimensional embeddings.

% \todo{can u make a nice figure showing the multi-task prediction and how it integrates with the batch estimator}
% The core steps of our estimator involve training discriminators $D_1$, $D_2$, and $D_{12}$ to estimate the conditional probabilities $p(y|x_1), p(y|x_2)$, and $p(y|x_1, x_2)$, followed by aligning the marginal distributions of $q$ and $p$. In the multi-scale BATCH estimator, the discriminators are lag-conditioned and share a common backbone encoder across all lags, while producing outputs for each lag through lag-specific heads. The alignment of $q$ is likewise conditioned on the corresponding time lag. Further details can be found in Appendix \ref{sec:batch}.
% \todo{batch and router figs can be together}

% \todo{provide formal definition of dynamics - would be good to have three examples of dynamics: temporal dynamics so u need to model interactions over time in 2 sequences, spatial dynamics so u need to model interactions over space like different regions of an image/video, spatio-temporal dynamics where one modality has elements in a temporal sequence and the other has spatial elements in a image etc.}

\subsection{\name: Building RUS-Aware MoE Routers}
\label{sec:moe}

We now discuss how to link routing decisions with information-theoretic principles to make the MoE interaction-aware. We begin by outlining the routing strategies for different interaction types. We then present the RUS-aware router, which incorporates the context of RUS sequences into MoE. Finally, we describe the auxiliary loss that enforces these routing strategies during training.

% \todo{give a high level overview of the parts.. first we need to understand how the information links to models.. then how we train and how we route.. }

% Prior work has shown the promise of using data quantification to select a suitable multimodal fusion architecture~\citep{liang2023quantifying,liang2024multimodal} [TODO cite more papers], without accounting for the temporal dimension of multimodal data. The key impact of our temporal RUS framework is to enable fine-grained multimodal fusion experts across time. For example, unimodal models could be prioritized at the beginning if there is only unique intformation present, and cross-attention fusion could be prioritized at certain time period when there is more display of synergy. Furthermore, it could identify which asynchronized time periods would multimodal fusion be most critical, instead of naively and inefficiently perform fusion across all possible time steps. This will allow us to develop multimodal models that are both performant and efficient.

\begin{table}[t]
\centering
\begin{tabularx}{0.98\textwidth}{c|c|c|c}
\Xhline{4\arrayrulewidth}
\textbf{RUS Values} & \textbf{Routing Strategies} & \textbf{Expert Types} & \textbf{Equivalent Fusion Types} \\
\hline
High $R_{m_1, m_2}$ & Route tokens together & Regular Expert (shared) & Early Fusion \\
%\hline
High $U_{m_1}$ & Diversify routing & Regular Expert & Late Fusion \\
%\hline
High $S_{m_1, m_2}$ & Route tokens together & Cross-Modal Expert & Hybrid Fusion \\
\Xhline{4\arrayrulewidth}
\end{tabularx}
\caption{Summary of our proposed routing strategies that leverage temporal RUS insights. Such a mechanism can also be viewed as a mixture of different fusion techniques \citep{baltruvsaitis2018multimodal}.}
\label{tab:rus_strategies}
\end{table}

\textbf{Routing strategies.} Our model is based on the principle that different types of interactions call for different computational strategies: (1) When multiple modalities contain similar information (\textbf{R}), they can be routed to the same expert. This is supported by both theoretical insights from Kolmogorov complexity \citep{rissanen1978modeling} and empirical findings in multimodal learning \citep{baltruvsaitis2018multimodal, ngiam2011multimodal}. (2) When modalities contain distinct information (\textbf{U}), they should be routed to different experts to fully capture their uniqueness. Such routing diversification has also been highlighted in recent studies \citep{mustafa2022multimodal, han2024fusemoe} as a means of ensuring balanced expert utilization. (3) Since high synergy (\textbf{S}) benefits from explicit interaction modeling \citep{liang2023quantifying, liu2018efficient}, we design dedicated synergy experts to capture cross-modal interactions. Each synergy expert consists of a cross-attention module followed by a feed-forward layer. In Table~\ref{tab:rus_strategies}, we summarize the routing strategy and corresponding fusion techniques.

\begin{wrapfigure}[12]{r}{0.49\textwidth}
\vspace{-1.8em}
\begin{center}
    \includegraphics[width=.5\textwidth]{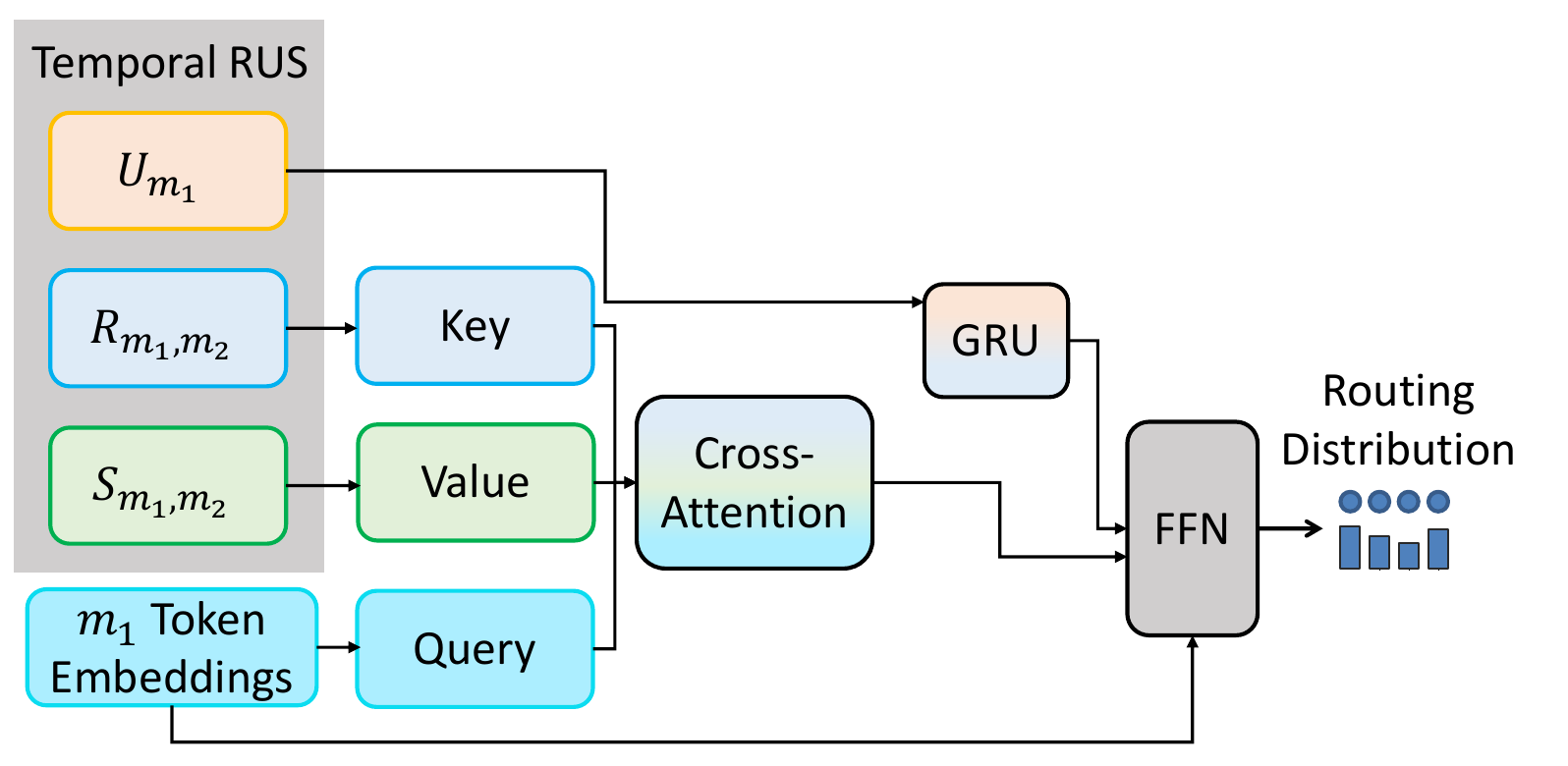}
\end{center}
\vspace{-1.2em}
\caption{\footnotesize{Router structure: routing decisions weighs RUS-derived features and token embeddings.}}
\label{fig:rus_router}
\end{wrapfigure}

\textbf{RUS-aware router} is the central component that incorporates temporal RUS to guide MoE training. As illustrated in Figure \ref{fig:rus_router}, the attention mechanism focuses on pairwise redundancy and synergy between modalities, which weights the importance of different pairwise interactions based on the current token's characteristics. A GRU-based module captures the temporal dynamics of uniqueness, it further combines with attention-weighted interaction contexts. The final routing decision fuses token-specific representations with RUS-derived features. Formally, for each modality $m_1 \in \{1, 2, \ldots, M\}$ and all other modalities $m_2 \neq m_1$, let $\oplus$ denote concatenation. The RUS-aware router is formalized as
\begin{align}
\text{RUSContext}_{m_1} &= \text{Attention}\left(\text{Query}_{m_1}, \{[R_{m_1,m_2}, S_{m_1,m_2}]\}_{m_1 \neq m_2}\right) \oplus \text{GRU}(U_{m_1}),
\label{eq:ruscontext} \\
\text{Logits}_{m_1} &= \text{MLP}\left(\text{TokenFeatures}_{m_1} \oplus \text{RUSContext}_{m_1}\right).
\label{eq:mlp_combine}
\end{align}
This design ensures that routing decisions for modality $m_1$ are informed by its temporal interactions with all other modalities $m_2$ (captured by temporal RUS), not just its individual content.

\textbf{Auxiliary losses for multimodal interaction experts.} The training objective is formulated to ensure that the router adheres to the predefined principles. It combines the task-specific loss with a RUS-aware auxiliary loss, where the latter decomposes into components aligned with each interaction type. For redundancy, at each time step $t$, we employ Jensen–Shannon Divergence (JSD) to quantify the discrepancy between the routing distributions $P_{\text{router}}$ of $m_1$ and $m_2$. As shown in Eq. \ref{eq:r_loss}, minimizing $\mathcal{L}_{\text{redundancy}}$ encourages $m_1$ and $m_2$ to be routed to the same expert whenever their redundancy exceeds the threshold $\tau_R$. 
\begin{align}
\mathcal{L}_{\text{redundancy}} &= \lambda_R \cdot \frac{1}{N_{\text{redundant}}} \sum_{(m_1, m_2, t): R_{m_1,m_2,t} > \tau_R} \text{JSD}(P_{\text{router}}^{(t,m_1)}, P_{\text{router}}^{(t,m_2)}).
\label{eq:r_loss}
\end{align}
Similarly, for uniqueness, as shown in Eq. \ref{eq:u_loss}, we encourage diversified routing of $m_1$ and $m_2$ whenever their uniqueness exceeds the threshold $\tau_U$ (uniqueness is modality-specific, not pairwise).
\begin{equation}
\mathcal{L}_{\text{uniqueness}} = -\lambda_U \cdot \frac{1}{N_{\text{unique}}} \sum_{(m_1, m_2, t): U_{m_1,t} > \tau_R \cap U_{m_2,t} > \tau_R} \text{JSD}(P_{\text{router}}^{(t,m_1)}, P_{\text{router}}^{(t,m_2)}).
\label{eq:u_loss}
\end{equation}
We promote routing to synergy experts whenever synergistic information is high:
\begin{equation}
\mathcal{L}_{\text{synergy}} = \lambda_S \cdot \frac{1}{N_{\text{synergy}}} \sum_{(m_1, m_2, t): S_{m_1,m_2,t} > \tau_S} \left(1 - \frac{P_{\text{syn}}^{(t,m_1)} + P_{\text{syn}}^{(t,m_2)}}{2}\right).
\label{eq:s_loss}
\end{equation}
The overall objective combines the task-specific loss and the interaction-aware losses defined in Eq. \ref{eq:r_loss}–\ref{eq:s_loss}. Importantly, the train–test split used for temporal RUS computation is aligned with that of \name’s training, ensuring that the estimated interactions directly facilitate the learning of the corresponding tokens. Note that, the RUS estimation is performed separately from the \name~training phase. While it is possible to build an end-to-end framework that jointly learns temporal RUS values and MoE model, we intentionally decouple these two components for the following reasons: (1) the multi-scale batch estimator is designed to reflect intrinsic information-theoretic structure of the dataset. It should not be optimized to minimize task-specific losses, doing so would entangle the RUS values with downstream objectives, undermining their role as task-agnostic signals of temporal multimodal interactions. (2) Since temporal RUS is a property of the dataset, it only needs to be computed \textit{once}. The resulting values can then be cached and reused across any number of downstream tasks for this dataset. We expect \name ~to leverage these implicit interaction dynamics, thereby enhancing both the performance and the interpretability of downstream tasks.
% \todo{write out the final training objective loss + loss redundacy + ...} \todo{and explain how hyperparams are tuned to balance the 4? loss terms}

\section{Experiments} \label{sec:exp}

% \paul{again, be consistent with terminology whether temporal RUS or temporal interactions etc}

\textbf{Overview.} We conduct comprehensive empirical evaluations of \name, aiming to answer the following questions: (1) What insights can temporal RUS provide in different scenarios? (2) Can leveraging temporal RUS improve MoE training accuracy? (3) How do temporal RUS influence task outcomes? (4) How can we interpret the learned routing assignments? Our evaluation spans diverse domains, including healthcare, activity recognition, and affective computing. In these evaluations, we incorporate a wide range of heterogeneous input streams, including sensors, wearables, medical imaging, text, and ECG, as distinct modalities to demonstrate \name’s ability to capture complex temporal interactions among them and its effectiveness in massively multimodal settings.

% firstly, what is the accuracy of identifying the temporal multimodal interactions? and are we doing so efficiently? are the temporal interactions 

% secondly, does our mixture of experts model selected by the temporal multimodal interactions perform better than full model fusion... and other temporal models... and more efficiently...?

\begin{table}[t]
\centering
\renewcommand\arraystretch{2.1}
\caption{\name~demonstrates superior results across different benchmarks and datasets. The best results are highlighted in \textbf{bold font}, and the second-best results are \underline{underlined}.}
\label{tab:model_performance}
% Resize the table to fit within the text width
% \scalebox{0.47}{
\setlength{\tabcolsep}{3pt} % tighter columns
\begin{adjustbox}{max width=\textwidth}
% \resizebox{1.05\textwidth}{!}{%
% \setlength{\tabcolsep}{3pt}{ % reduce column spacing
% \fontsize{14pt}{12pt}\selectfont
\hspace{-1ex}
\begin{tabular}{l|cc|cc|cc|cc|cc|c}
\Xhline{8\arrayrulewidth}
\textbf{\Large Datasets} & \multicolumn{2}{c|}{\textbf{\Large PAMAP2}} & \multicolumn{2}{c|}{\textbf{\Large MIMIC-IV IHM}} & \multicolumn{2}{c|}{\textbf{\Large MIMIC-IV LOS}} & \multicolumn{2}{c|}{\textbf{\Large MOSI}} & \multicolumn{2}{c|}{\textbf{\Large WESAD}} & \textbf{\Large Opportunity} \\ \hline
\textbf{\Large Metrics} & \textbf{\Large Accuracy} & \textbf{\Large F1} & \textbf{\Large AUROC} & \textbf{\Large F1} & \textbf{\Large AUROC} & \textbf{\Large F1} & \textbf{\Large Accuracy} & \textbf{\Large AUROC} & \textbf{\Large Accuracy} & \textbf{\Large AUROC} & \textbf{\Large Accuracy} \\ \hline
\Large Transformer & {\Large 82.48} $\pm$ 1.26 & {\Large 82.57} $\pm$ 1.52 & {\Large 80.18} $\pm$ 0.17 & {\Large 78.96} $\pm$ 0.65 & {\Large 75.46} $\pm$ 0.46 & {\Large 72.31} $\pm$ 0.56 & {\Large 68.39} $\pm$ 0.36 & {\Large 68.47} $\pm$ 0.41 & {\Large 53.84} $\pm$ 1.27 & {\Large 74.39} $\pm$ 1.15 & {\Large 81.59} $\pm$ 0.74 \\
\Large mTAND       & {\Large 74.62} $\pm$ 0.53 & {\Large 74.38} $\pm$ 0.75 & {\Large 80.89} $\pm$ 0.33 & {\Large 79.35} $\pm$ 0.49 & {\Large 77.34} $\pm$ 0.29 & {\Large 73.45} $\pm$ 0.44 & {\Large 70.07} $\pm$ 0.47 & {\Large 69.94} $\pm$ 0.29 & {\Large 48.22} $\pm$ 0.88 & {\Large 71.66} $\pm$ 0.97 & {\Large 70.26} $\pm$ 2.03 \\
\Large MulT        & {\Large 82.23} $\pm$ 0.39 & {\Large 81.87} $\pm$ 0.43 & {\Large 81.63} $\pm$ 0.47 & {\Large 81.55} $\pm$ 0.53 & {\Large 76.68} $\pm$ 0.93 & {\Large 72.52} $\pm$ 0.77 & {\Large 68.80} $\pm$ 0.78 & {\Large 69.05} $\pm$ 0.83 & {\Large 47.63} $\pm$ 0.43 & {\Large 71.43} $\pm$ 0.33 & {\Large 72.61} $\pm$ 0.89 \\
\Large MISTS       & {\Large 85.34} $\pm$ 0.78 & {\Large 85.79} $\pm$ 0.31 & {\Large 82.21} $\pm$ 0.75 & {\Large 80.56} $\pm$ 0.72 & {\Large 77.49} $\pm$ 0.65 & {\Large 73.86} $\pm$ 0.35 & {\Large 69.42} $\pm$ 0.32 & {\Large 69.32} $\pm$ 0.51 & {\Large 51.72} $\pm$ 0.71 & {\Large 73.29} $\pm$ 0.61 & {\Large 79.36} $\pm$ 1.41 \\
\Large FuseMoE     & \underline{{\Large 87.74} $\pm$ 0.49} & \underline{{\Large 86.51} $\pm$ 1.17} & {\Large 82.33} $\pm$ 0.45 & {\Large 81.64} $\pm$ 0.58 & \underline{{\Large 81.74} $\pm$ 0.65} & \textbf{{\Large 75.18} $\pm$ 0.41} & \textbf{{\Large 75.65} $\pm$ 1.56} & \textbf{{\Large 78.43} $\pm$ 0.97} & \underline{{\Large 53.92} $\pm$ 1.14} & \underline{{\Large 76.31} $\pm$ 0.99} & \underline{{\Large 83.15} $\pm$ 1.12} \\
\Large I2MoE       & {\Large 84.55} $\pm$ 0.64 & {\Large 84.24} $\pm$ 0.33 & \underline{{\Large 83.28} $\pm$ 0.27} & \underline{{\Large 82.59} $\pm$ 0.30} & {\Large 79.88} $\pm$ 1.08 & {\Large 74.36} $\pm$ 0.89 & {\Large 71.91} $\pm$ 2.20 & {\Large 74.88} $\pm$ 1.33 & {\Large 52.64} $\pm$ 1.06 & {\Large 75.52} $\pm$ 1.34 & {\Large 82.16} $\pm$ 2.57 \\ \hline
\cellcolor{ashgrey} \Large \name    & \textbf{{\Large 91.37} $\pm$ 1.38} & \textbf{{\Large 90.44} $\pm$ 1.02} & \textbf{{\Large 85.40} $\pm$ 0.24} & \textbf{{\Large 84.97} $\pm$ 0.35} & \textbf{{\Large 81.88} $\pm$ 0.18} & \underline{{\Large 74.43} $\pm$ 0.46} & \underline{{\Large 72.04} $\pm$ 1.99} & \underline{{\Large 77.25} $\pm$ 0.91} & \textbf{{\Large 55.74} $\pm$ 1.99} & \textbf{{\Large 77.34} $\pm$ 1.00} & \textbf{{\Large 84.32} $\pm$ 1.33} \\ \Xhline{8\arrayrulewidth}
% \end{tabular}}
\end{tabular}
\end{adjustbox}
\end{table}

\subsection{Experiment Setup}
\textbf{Dataset information.} ~
\textbf{PAMAP2} \citep{reiss2012introducing} is a physical activity monitoring benchmark that includes recordings of 18 activities, carried out by 9 participants equipped with 3 inertial measurement units and a heart rate monitor. The \textbf{MIMIC-IV} \citep{johnson2023mimic} ecosystem contains deidentified clinical data from ICU patients, including lab measurements, vitals, notes, and chest X-rays. We consider two tasks: in-hospital mortality (IHM) and length-of-stay (LOS) prediction. \textbf{CMU-MOSI} \citep{zadeh2016mosi} is a multimodal dataset of 2,199 annotated YouTube vlog opinion segments with audio, video, and text, used here for binary sentiment classification. \textbf{WESAD} \citep{schmidt2018introducing} is a wearable sensor dataset from 15 subjects in a lab study, with physiological and motion data from chest- and wrist-worn devices. It supports affect recognition across six states, including stress, amusement, and meditation. %\textbf{OPPORTUNITY} is a human activity recognition dataset, capturing wearable, object, and ambient sensors.
\textbf{Opportunity} \citep{chavarriaga2013opportunity} provides multi-modal sensor recordings from wearable, object, and ambient devices, capturing natural and scripted daily activities of 4 subjects in a realistic home-like environment.

\textbf{Baselines.}
To validate the effectiveness of \name~in enhancing multimodal learning, we benchmark it against a diverse set of state-of-the-art models. The comparison covers: (1) standard monolithic backbones, including multi-head self-attention based Transformers \citep{vaswani2017attention} and multi-time attention for irregular time series (mTAND) \citep{shukla2021multi}; (2) task-specific fusion models, such as MulT \citep{tsai2019multimodal} and MISTS \citep{zhang2023improving}; and (3) alternative multimodal MoE approaches, including FuseMoE \citep{han2024fusemoe} that does not incorporate multimodal interaction, and I2MoE \citep{xin2025i2moe} that uses interaction-specific experts.

\textbf{Implementation.}
We split PAMAP2, WESAD, and Opportunity into training, validation, and test sets by subject. For MIMIC-IV, we use the first 48 hours of each ICU stay as input and randomly assign the stays to training, validation, or test splits. For MOSI, we adopt the official train/val/test splits provided in Multibench \citep{liang2021multibench}. Unlike the other datasets, MIMIC-IV and MOSI contain many more sequences, and the target for each sequence is a single static label rather than a label sequence. When computing RUS values for these datasets, we focus on a single time step $n$, taking $Y^n$ as the label for each input sequence while still considering multiple lags $\tau$ from the end of the sequence. Because the RUS computation aggregates across multiple sequences, the resulting variety of labels ensures that the estimation of $P(Y^n)$ does not produce a degenerate distribution.
% In outcome-based studies, such as clinical applications, the outcome $Y$ is normally a static label rather than a temporal sequence. In such cases, multiple samples are required to form valid distributions, since a single instance would result in a degenerate distribution for $Y$.
\begin{figure}[t]
    \begin{minipage}{0.95\textwidth}
    \centering
    \begin{tabular}{@{\hspace{-2.4ex}} c @{\hspace{-0.5ex}} @{\hspace{-2.4ex}} c @{\hspace{-0.5ex}} @{\hspace{-2.4ex}} c @{\hspace{-0.5ex}} @{\hspace{-2.4ex}} c @{\hspace{-2.4ex}}}
        \begin{tabular}{c}
        \includegraphics[width=.27\textwidth]{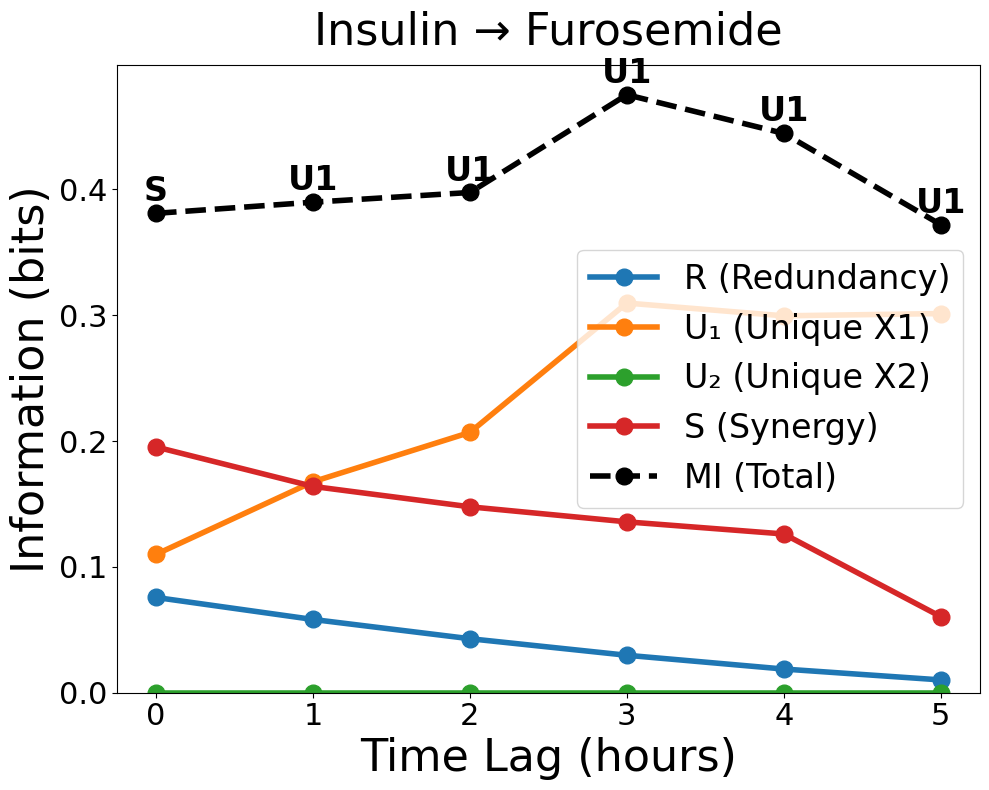}
        %\vspace{-5pt}
        \\
        {\small{(a) MIMIC-IV}}
        \end{tabular} &
        \begin{tabular}{c}
        \includegraphics[width=.27\textwidth]{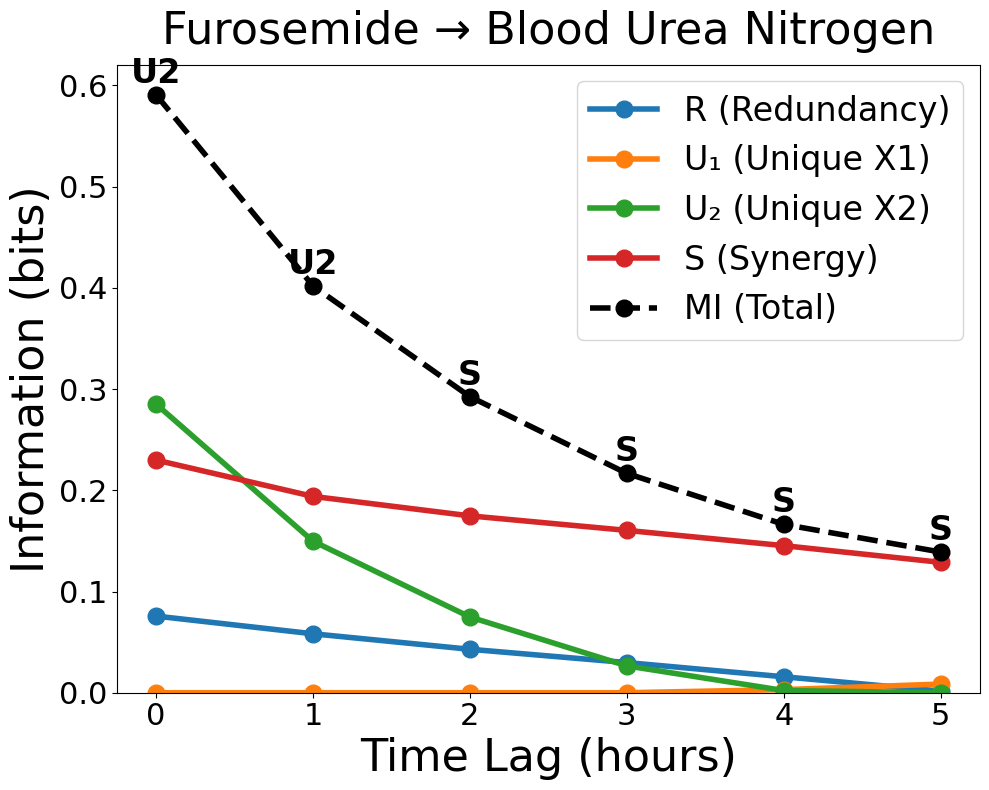}
        %\vspace{-5pt}
        \\
        {\small{(b) MIMIC-IV}}
        \end{tabular} & 
        \begin{tabular}{c}
        \includegraphics[width=.27\textwidth]{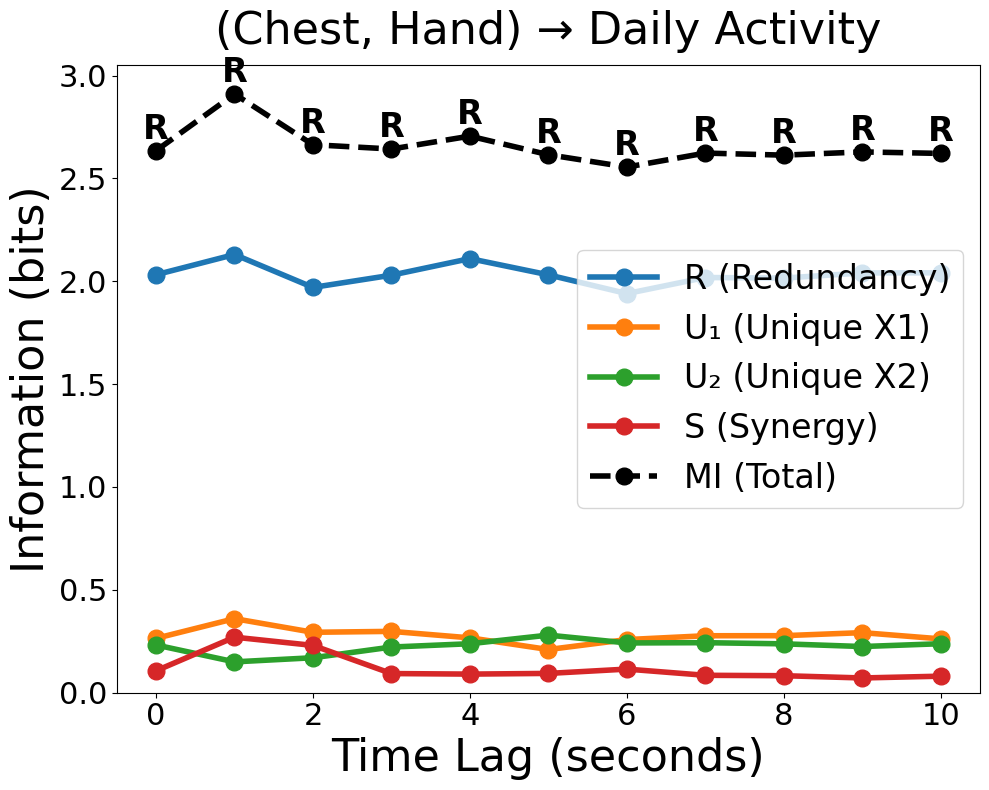} 
        %\vspace{-5pt}
        \\
        {\small{(c) PAMAP2}}
        \end{tabular} &
        \begin{tabular}{c}
        \includegraphics[width=.27\textwidth]{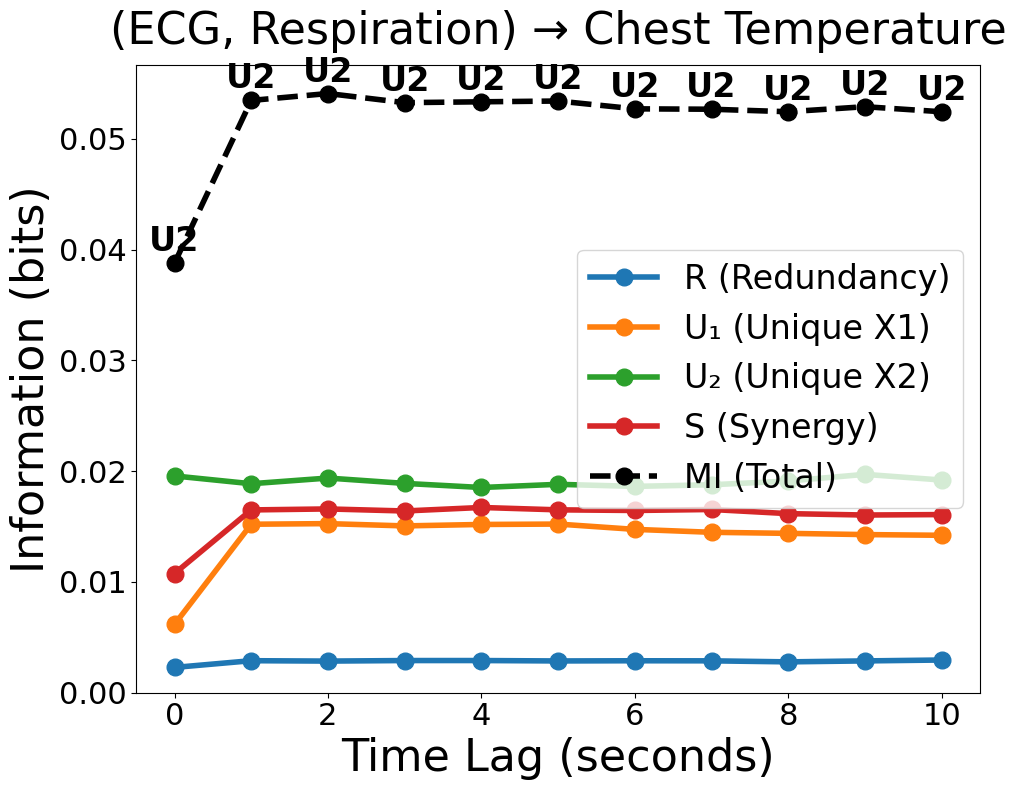}
        %\vspace{-5pt}
        \\
        {\small{(d) WESAD}}
        \end{tabular} \\
        \end{tabular}
    \end{minipage}
    % \vspace{-0em}
    \caption{Insights from temporal RUS values across different applications include: (a) insulin and furosemide exhibit a strong synergistic effect at the time of administration, while insulin’s unique effect becomes more pronounced later; (b) as time progresses after furosemide administration, its physiological impact increases; (c) in activity recognition, chest and hand motion display coupled movements during locomotion, reflecting strong redundancy; and (d) in physiological monitoring, ECG and respiration signals from one second prior provide better predictions of current chest temperature, capturing the natural response delay to stimuli.}
    \label{fig:trus_insights}
\end{figure}

\subsection{Performance and Insights of \name}

% \todo{where is the explanation and summary of results first? before talking insights?}
\textbf{\name\ demonstrates superior performance.}~
In Table \ref{tab:model_performance}, we compare the performance of \name~against various baseline models across 6 multimodal benchmark tasks. Results (averaged over 5 random runs) demonstrate that \name~consistently achieves superior performance, attaining the best results in nearly all metrics. In particular, for healthcare (MIMIC-IV) and affective computing tasks (MOSI and WESAD), \name~outperforms task-specific fusion methods such as MulT and MISTS by a significant margin. Against multimodal MoE baselines including FuseMoE and I2MoE, \name~also achieves improved results. These findings highlight that leveraging interaction dynamics across modalities can effectively enhance model performance, and that \name~generalizes well across diverse multimodal learning tasks.

\textbf{Qualitative examples of temporal RUS.}~
To understand the reason behind the improved performance, we examine the temporal RUS values obtained across different tasks to assess whether they provide rich and interpretable insights. For example, the changing relationship between insulin and furosemide over time in MIMIC-IV, shown in Fig.\ref{fig:trus_insights} (a), illustrates the interplay between their immediate and delayed physiological effects. At the time of simultaneous administration, a synergistic effect emerges, likely driven by their concurrent but opposing influences on blood glucose and potassium levels. However, when insulin is administered 1 hour earlier or more, its uniqueness becomes much stronger, reflecting its relatively rapid onset of action and peak effect occurring within 2–3 hours. As shown in Fig.\ref{fig:trus_insights} (b), at the time of Furosemide administration, its immediate diuretic effect is unique to the drug and is not yet fully captured in the Blood Urea Nitrogen (BUN) levels. Over time, however, the synergy between Furosemide and BUN increases, reflecting the delayed manifestation of the drug’s physiological effects in blood chemistry.

For activity recognition on PAMAP2, during activities such as walking, running, or climbing stairs, the natural swinging of the arms is directly coupled with the torso and chest movements. This relationship remains consistent over time, leading to a high redundancy pattern, as shown in Fig.~\ref{fig:trus_insights} (c). 
Lastly, examining the dynamics between ECG, respiration, and chest temperature in Fig.~\ref{fig:trus_insights} (d) reveals an initial increase in synergy and respiration uniqueness on WESAD dataset. This arises because the one-second delay aligns with the time it takes for the body’s stress response to begin manifesting as a change in skin temperature. After this brief adjustment period, the relationships stabilize, reflecting the steady state of the physiological response. These examples clearly demonstrate that temporal RUS can capture unique interaction dynamics for improved model performance.

\begin{figure}[t]
    \begin{minipage}{\textwidth}
    \centering
    % \vspace{-2ex}
    \begin{tabular}{@{\hspace{-3.8ex}} c @{\hspace{-2.4ex}} c @{\hspace{-1.5ex}} c @{\hspace{-1.5ex}}}
        \begin{tabular}{c}
        \includegraphics[width=.31\textwidth]{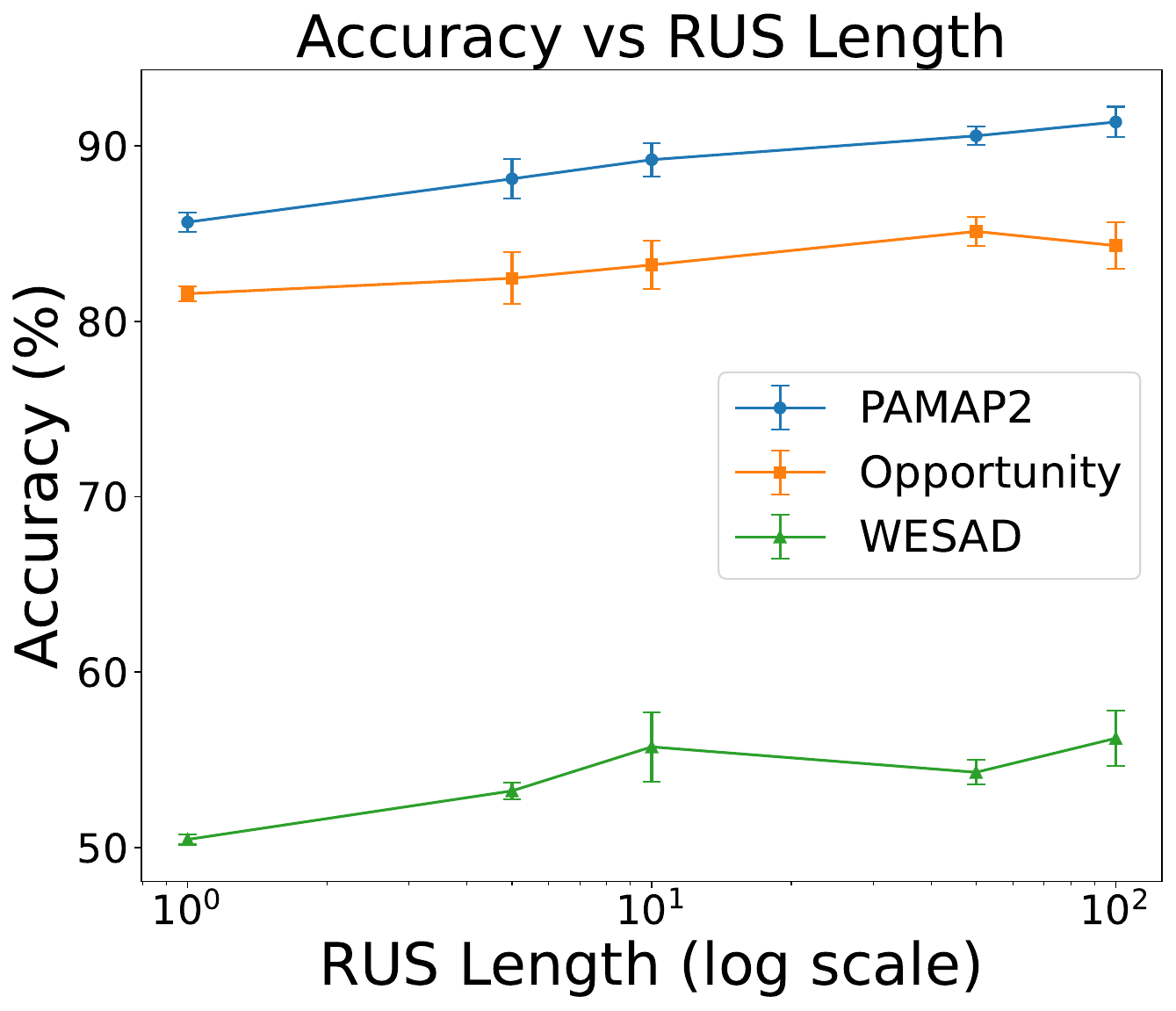}
        %\vspace{-5pt}
        \\
        {\small{(a)}}
        \end{tabular} & 
        \begin{tabular}{c}
        \includegraphics[width=.31\textwidth]{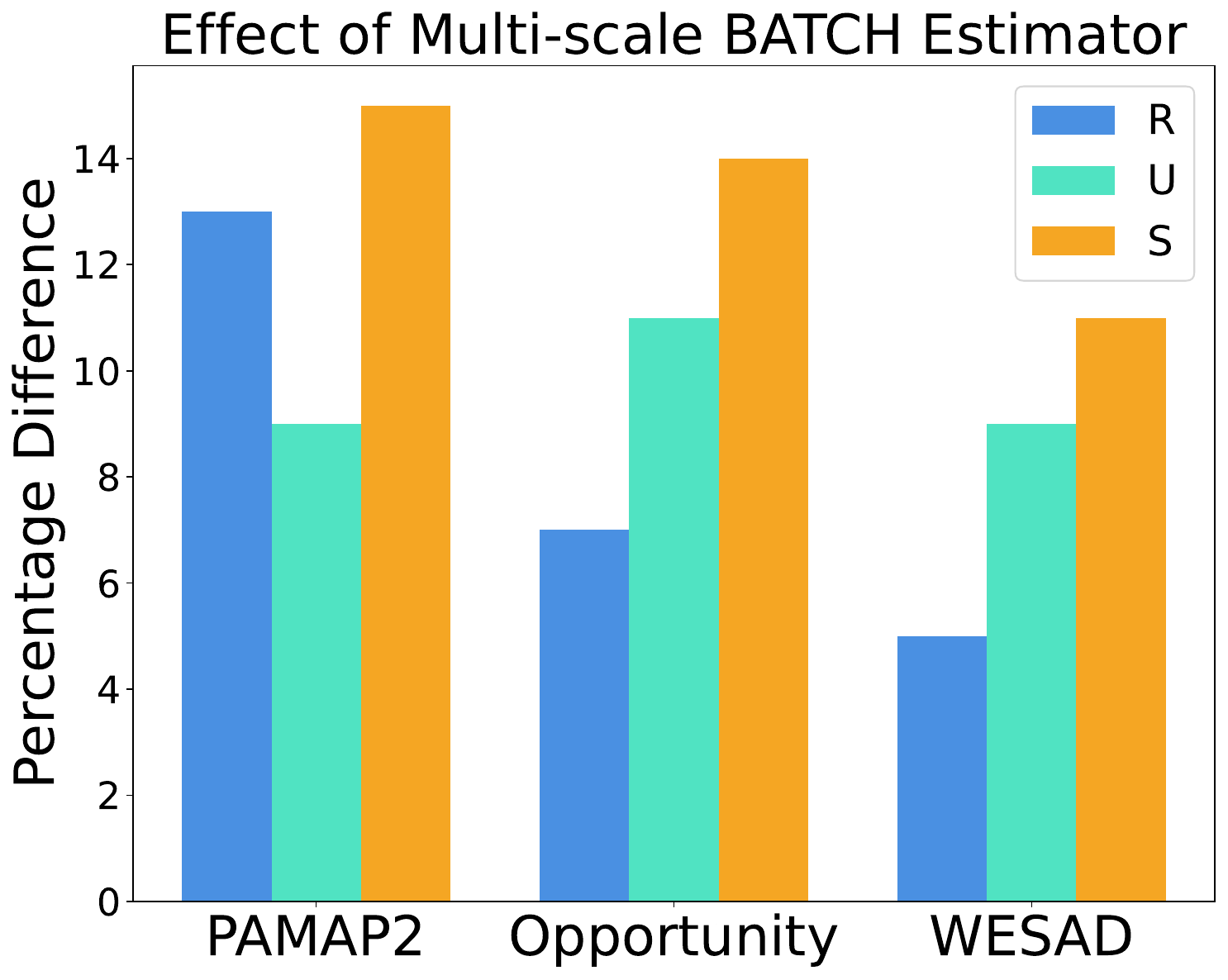} 
        %\vspace{-5pt}
        \\
        {\small{(b)}}
        \end{tabular} &
        \begin{tabular}{c}
        \includegraphics[width=.31\textwidth]{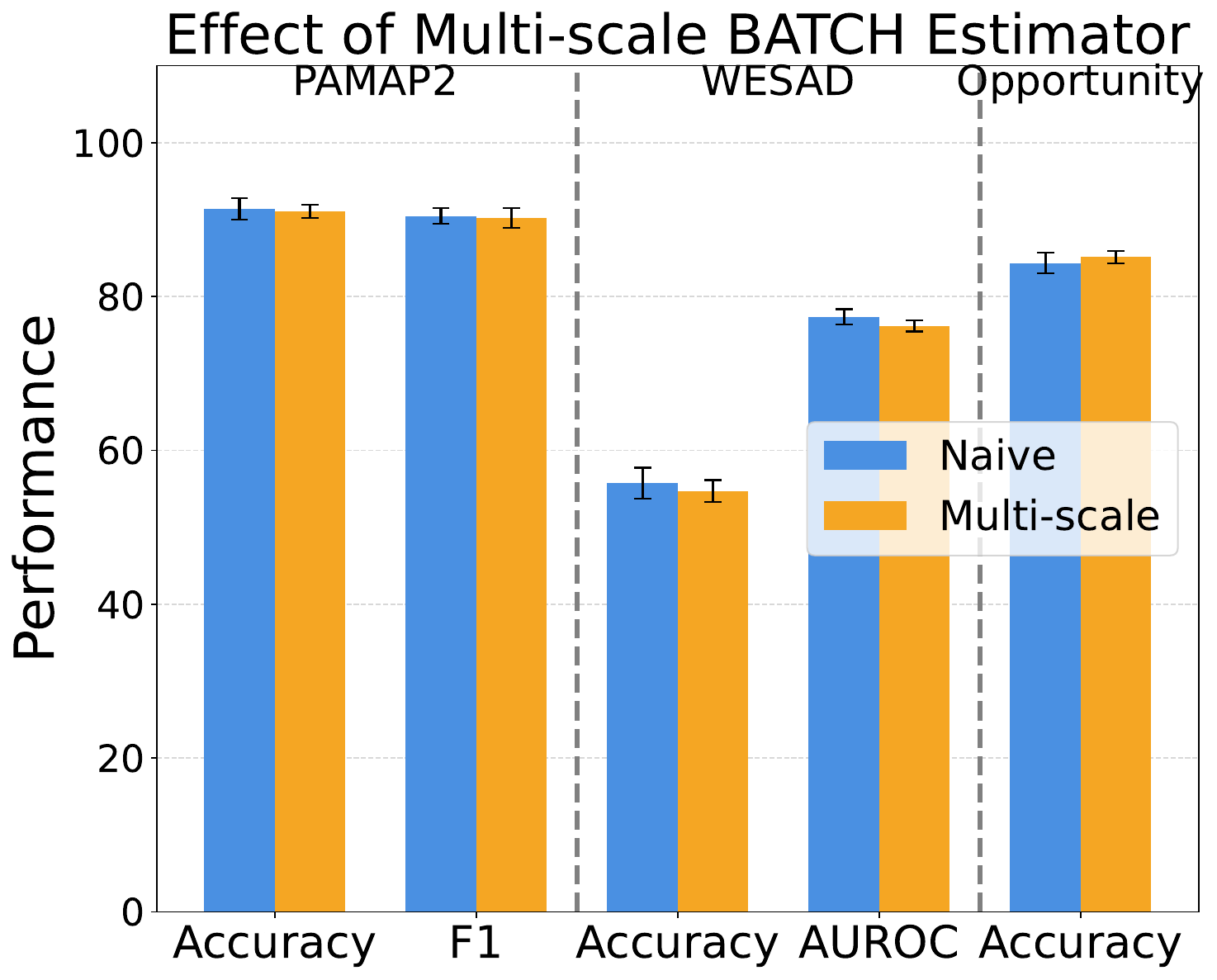} 
        %\vspace{-5pt}
        \\
        {\small{(c)}}
        \end{tabular} \\
        \end{tabular}
    \end{minipage}
    % \vspace{-1em}
    \caption{(a) Increasing temporal RUS length incorporated into MoE training improves performance; (b) discrepancy between the multi-scale BATCH estimator versus step-wise RUS computation; (c) performance differences across tasks after applying the multi-scale BATCH estimator.}
    \label{fig:rus_ablation}
\end{figure}

\begin{figure}[t]
    \begin{minipage}{0.95\textwidth}
    \centering
    \begin{tabular}{@{\hspace{-2.4ex}} c @{\hspace{-0.5ex}} @{\hspace{-2.4ex}} c @{\hspace{-0.5ex}} @{\hspace{-2.4ex}} c @{\hspace{-0.5ex}} @{\hspace{-2.4ex}} c @{\hspace{-2.4ex}}}
        \begin{tabular}{c}
        \includegraphics[width=.27\textwidth]{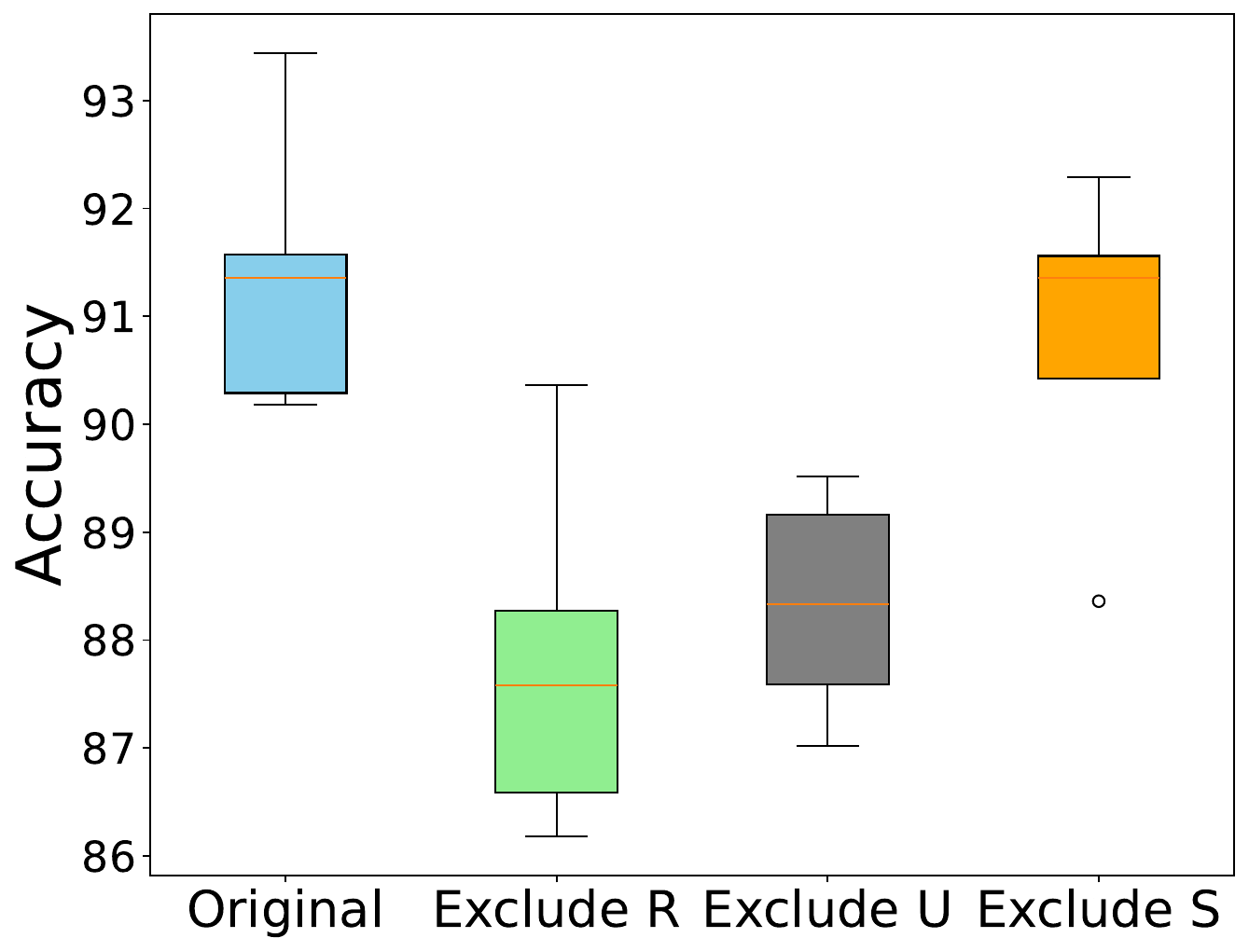}
        %\vspace{-5pt}
        \\
        {\small{(a) PAMAP2}}
        \end{tabular} &
        \begin{tabular}{c}
        \includegraphics[width=.27\textwidth]{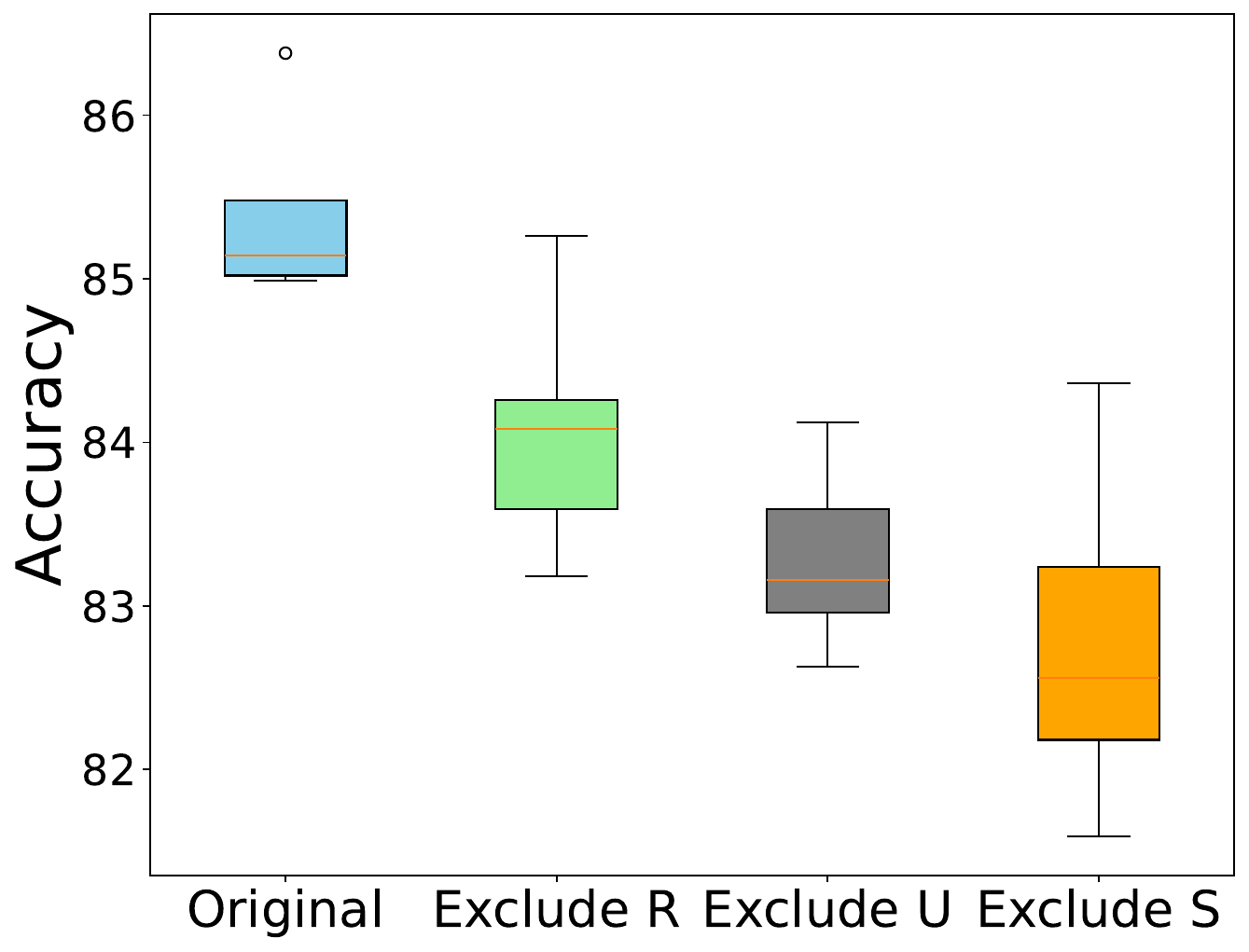}
        %\vspace{-5pt}
        \\
        {\small{(b) MIMIC-IV IHM}}
        \end{tabular} & 
        \begin{tabular}{c}
        \includegraphics[width=.27\textwidth]{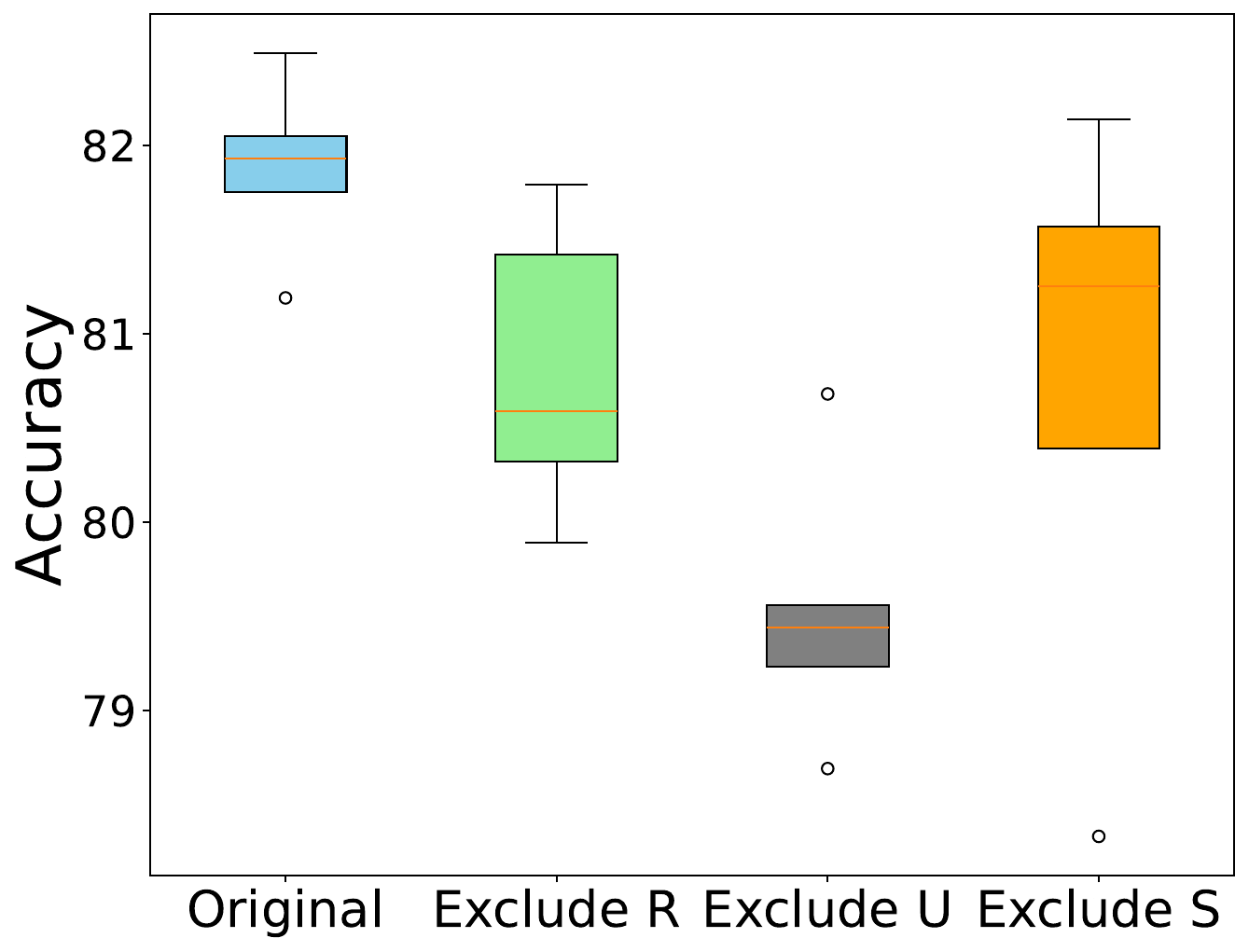} 
        %\vspace{-5pt}
        \\
        {\small{(c) MIMIC-IV LOS}}
        \end{tabular} &
        \begin{tabular}{c}
        \includegraphics[width=.27\textwidth]{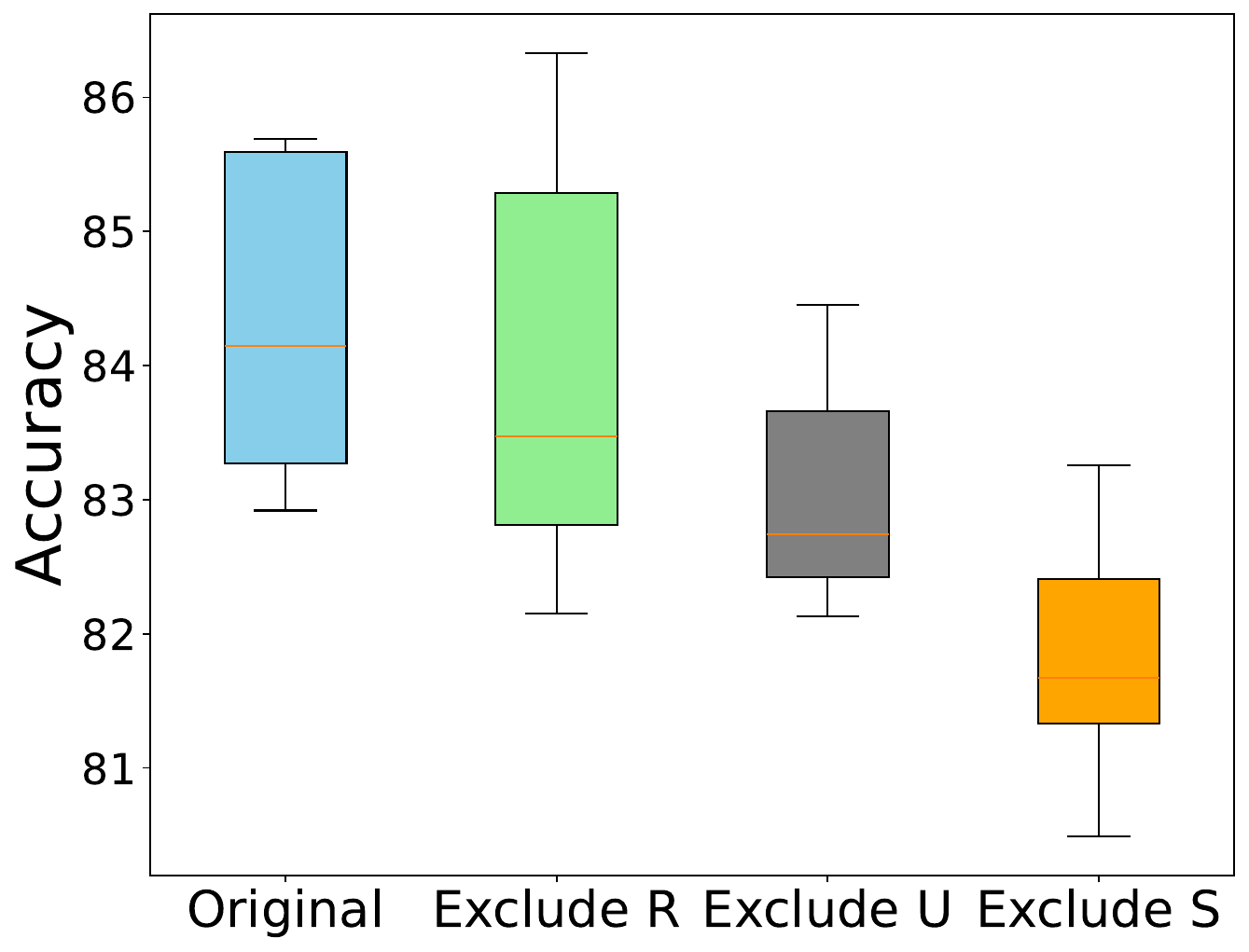}
        %\vspace{-5pt}
        \\
        {\small{(d) Opportunity}}
        \end{tabular} \\
        \end{tabular}
    \end{minipage}
    % \vspace{-1em}
    \caption{Performance change after removing each auxiliary loss term from Eq.\ref{eq:r_loss} to \ref{eq:s_loss}. The results show that all interactions contribute effectively to enhancing performance across nearly all settings.}
    \label{fig:rus_component}
\end{figure}

% We use the following datasets to evaluate the performance of out method. Datasets: priority should be datasets with very long sequence lengths from medical or multimedia domains. \todo{also emphasize tasks come from multimedia, healthcare, video analysis, affective computing and more}

% Other datasets: PAMAP, CMU-MOSI/MOSEI, IMDB, CLIMB

% \textbf{Other Implementation Details}~ train/val/test w.r.t. RUS, model sizes, complexity w.r.t. temporal RUS
\subsection{In-depth Analysis of \name}

\textbf{Impact of temporal RUS length.} We investigate how the length of temporal RUS incorporated into MoE training affects performance. Our intuition is that a longer temporal range captures richer interaction information, thereby improving results. We first increase the maximum time lag of the temporal RUS from 1 to 10. We do not extend it further, as estimating too many steps simultaneously can degrade the performance of the multi-scale BATCH estimator. To further extend sequence length, we repeat temporal RUS segments to form longer sequences. As shown in Fig.~\ref{fig:rus_ablation} (a), both strategies lead to performance gains. Increasing the maximum time lag provides a broader view of the temporal trajectory. Meanwhile, segment repetition appears to strengthen the model’s ability to recognize important modality interactions, improve context modeling, and reduce gradient variance. A similar observation that longer sequences with consistent patterns improve generalization has also been discussed in prior works \citep{mustafa2022multimodal, lai2018modeling}.
% longer RUS sequences lead to better performance, static is less good than temporal, 1 figure with 3 line plots (pamap, wesad, opportunity)

\textbf{Efficiency of multi-scale BATCH estimator.} The proposed multi-scale BATCH estimator achieves $\tau$-fold speedup in temporal RUS estimation while maintaining parameter efficiency. We compare the estimated RUS trajectories against the naive step-wise approach. As shown in Fig.\ref{fig:rus_ablation} (b), we report the percentage difference across tasks, averaged over all time lags. Although the magnitudes of individual interactions differ slightly, the discrepancies remain relatively small. Moreover, these differences do not lead to notable performance fluctuations, as demonstrated in Fig.\ref{fig:rus_ablation} (c).

\textbf{Impact of each auxiliary loss.} We further investigate the contribution of each auxiliary loss term to performance across tasks. To this end, we perform an ablation study by removing each auxiliary loss term from Eq.\ref{eq:r_loss} to \ref{eq:s_loss}, thereby disabling the corresponding routing mechanism. As shown in Fig.\ref{fig:rus_component}, uniqueness emerges as a critical factor for many tasks, while MIMIC-IV IHM and Opportunity are primarily synergy-driven, and PAMAP2 shows greater dependence on redundancy.

\begin{figure}[t]
    \begin{minipage}{0.95\textwidth}
    \centering
    \begin{tabular}{@{\hspace{-2.4ex}} c @{\hspace{-0.5ex}} @{\hspace{-2.4ex}} c @{\hspace{-0.5ex}} @{\hspace{-2.4ex}} c @{\hspace{-0.5ex}} @{\hspace{-2.4ex}} c @{\hspace{-2.4ex}}}
        \begin{tabular}{c}
        \includegraphics[width=.27\textwidth]{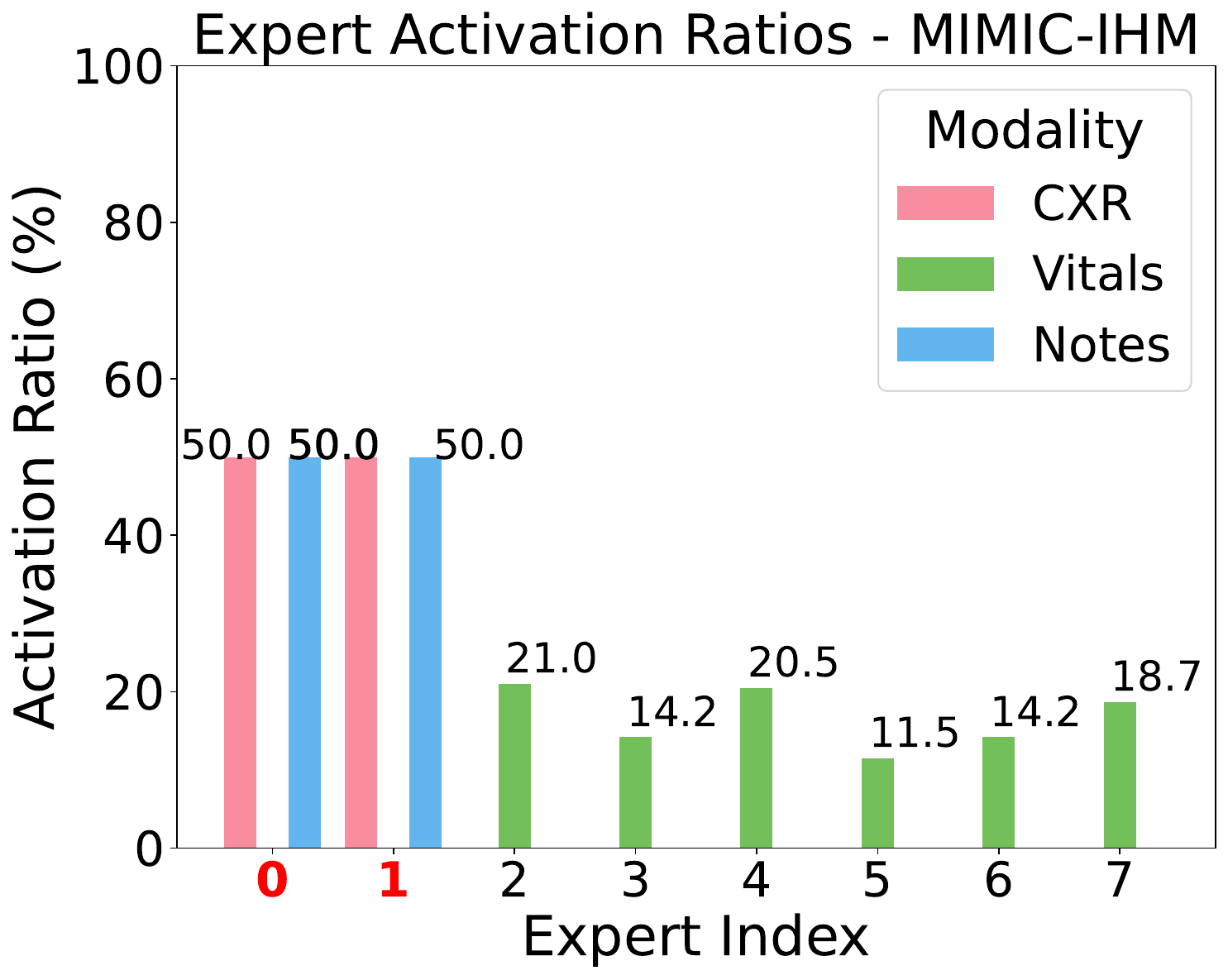}
        %\vspace{-5pt}
        \\
        {\small{(a) \name}}
        \end{tabular} &
        \begin{tabular}{c}
        \includegraphics[width=.27\textwidth]{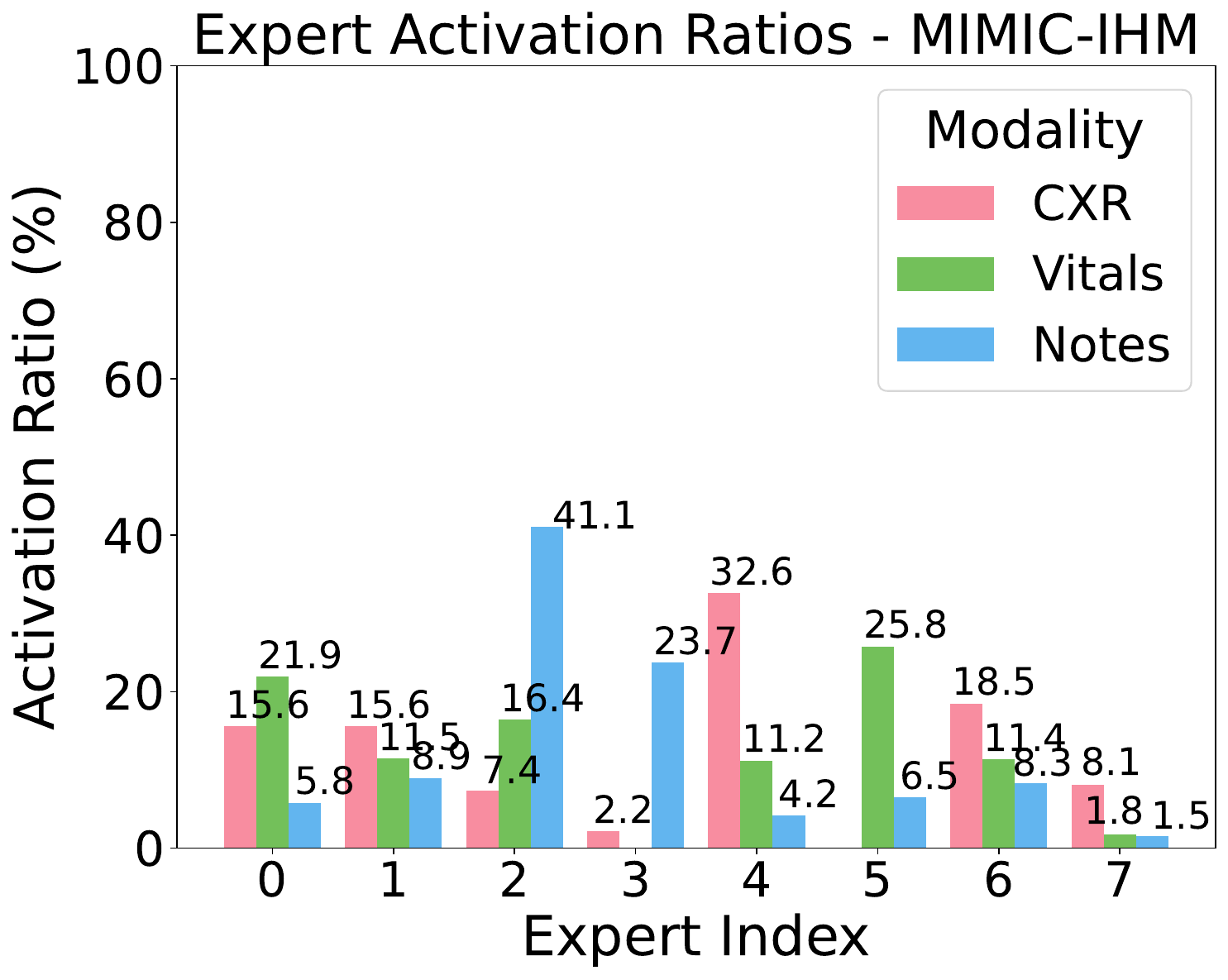}
        %\vspace{-5pt}
        \\
        {\small{(b) Standard MoE}}
        \end{tabular} & 
        \begin{tabular}{c}
        \includegraphics[width=.27\textwidth]{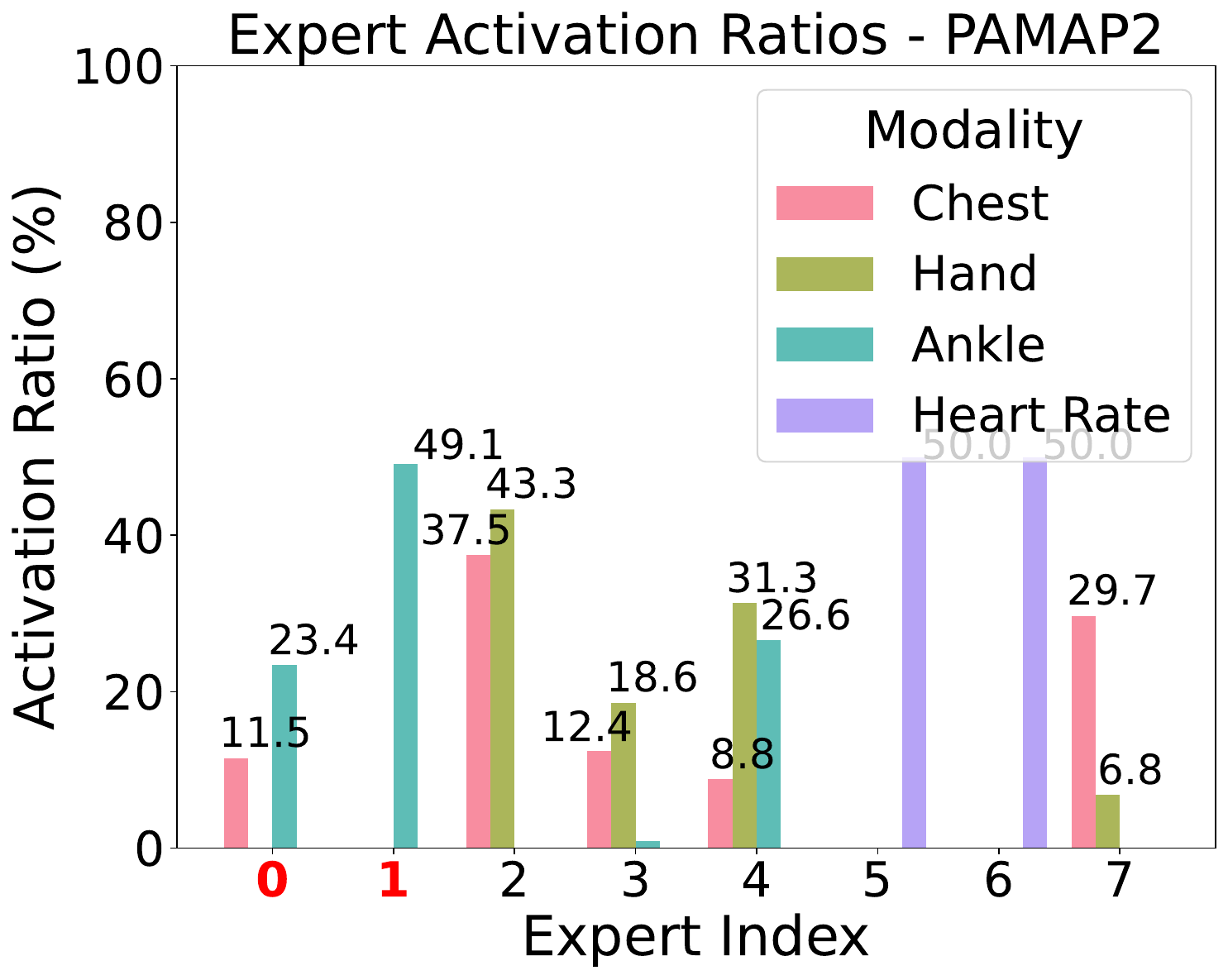} 
        %\vspace{-5pt}
        \\
        {\small{(c) \name}}
        \end{tabular} &
        \begin{tabular}{c}
        \includegraphics[width=.27\textwidth]{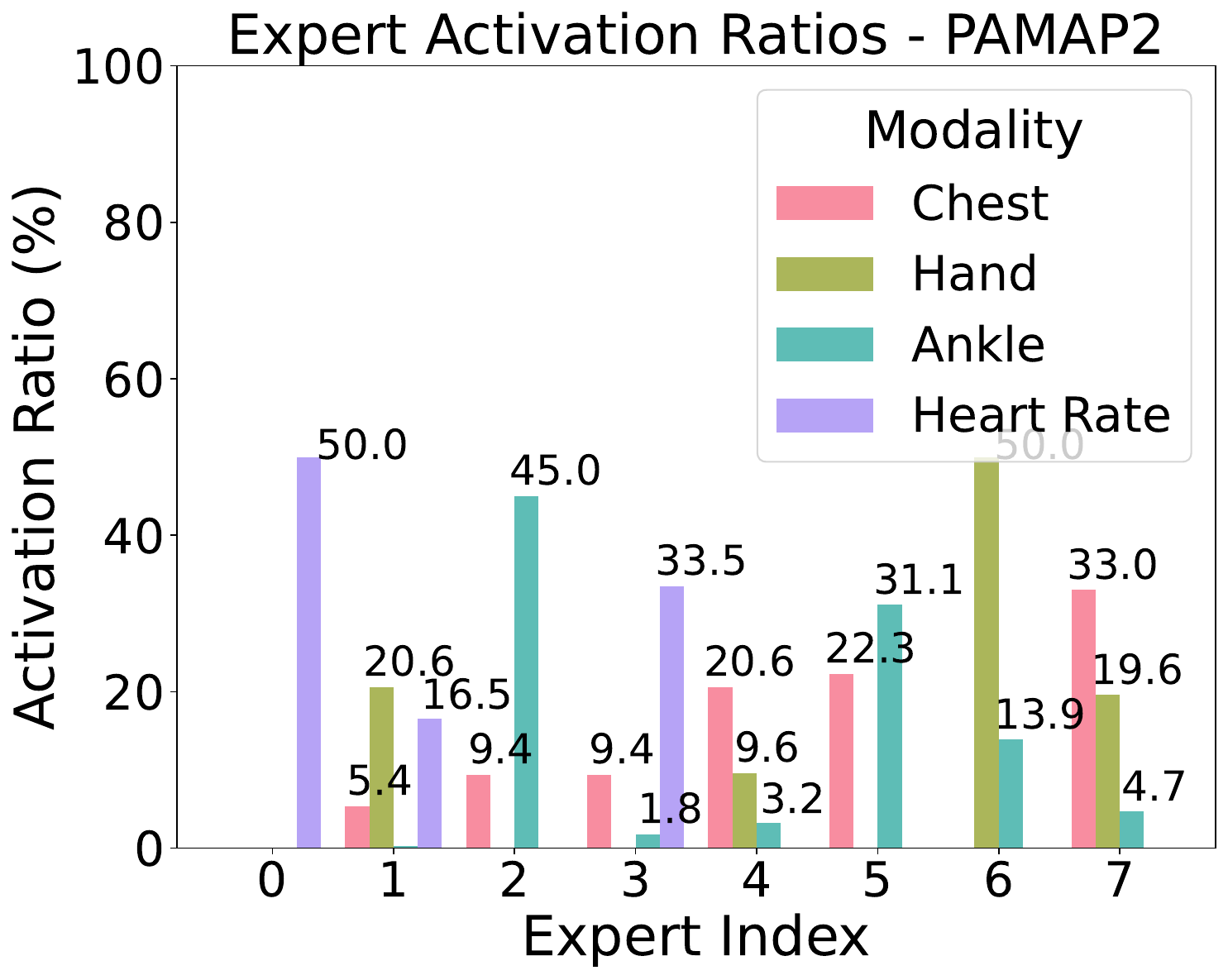}
        %\vspace{-5pt}
        \\
        {\small{(d) Standard MoE}}
        \end{tabular} \\
        \end{tabular}
    \end{minipage}
    % \vspace{-1em}
    \caption{Comparison of routing distributions between \name\ and standard MoE. The activation ratio is defined by the percentage of tokens from a specific modality that are routed to each expert. In (a) and (c), expert indices highlighted in red denote synergy experts. The plots show that \name\ adheres to the proposed routing principles and provides enhanced interpretability.}
    \label{fig:moe_routing}
\end{figure}

\textbf{Expert routing analysis.} Finally, we explicitly examine how \name\ influences multimodal token routing distributions. We compare \name\ with a standard MoE using top-$k$ routing, as shown in Fig.\ref{fig:moe_routing}. The distributions are generated from models that achieve the best performance for each task. In (a), RUS values show there is strong synergy between CXR and clinical notes, while vitals remain relatively independent. The router captures this by directing corresponding tokens to specialized experts, enabling them to be trained more effectively together. Likewise, in (c), chest and hand signals (with strong redundancy) are routed to the same expert, whereas modality pairs with high synergy or uniqueness are also handled according to their interaction type. In contrast, the standard MoE exhibits no such structured routing: experts process arbitrary combinations of modalities, resulting in unorganized assignments that hinder both efficient learning and interpretability.

\section{Conclusion and Future Work}
\label{sec:conclusion}

We introduced \name, a novel MoE architecture that integrates temporal multimodal interactions into model training. By decomposing multi-source directed information into temporal redundancy, uniqueness, and synergy, our approach captures rich evolving dynamics that effectively guide MoE routing in massively multimodal applications. We further developed scalable estimators suitable for multi-time-lag settings. \name~achieves substantial performance gains while preserving expert-level interpretability, making it a compelling design choice for future multimodal foundation models. Looking ahead, promising directions include extending the framework to capture more general spatio-temporal dynamics and investigating its applicability to multimodal and multitask learning scenarios. We would also like to extend the design philosophy of leveraging temporal multimodal interaction to guide expert routing to LLM-MoE frameworks. For instance, incorporating multimodal inputs with known temporal RUS values could potentially improve fine-tuning of MoE-based LLMs by helping the model identify the appropriate experts. Furthermore, we believe the current \name~design shows even greater potential when applied to large-scale VLMs or world-models, where temporal multimodal interaction dynamics remain highly informative and can meaningfully improve real-world performance.

% \newpage
\bibliography{reference}
\bibliographystyle{iclr2026_conference}

\newpage
\appendix
\section{Additional Related Works} \label{sec:trus_supp}
Partial Information Decomposition (PID) \citep{williams2010nonnegative, bertschinger2014quantifying} has emerged as a formal way to quantify multimodal interactions by measuring how the total information between two modalities $(x_1,x_2)$ useful for a task $y$ can be decomposed into redundant ($R$), unique ($U$), and synergistic ($S$) parts. Redundancy measures the common information between two modalities, uniqueness measures the useful information in a modality not present in the others, and synergy measures the new information that arises only when both modalities are fused. Specifically, given unimodal marginal distributions $p(x_1,y)$ and $p(x_2,y)$ over each modality and the multimodal joint distribution $p(x_1,x_2,y)$, a formal definition is
% \vspace{-3em}
\begin{align}
    R &= \max_{q\in \Delta_p} I_q(X_1;X_2;Y), \label{eqn:ri} \\
    U_1 &= \min_{q\in \Delta_p} I_q(X_1;Y|X_2),\quad U_2 = \min_{q\in \Delta_p} I_q(X_2;Y|X_1), \label{eqn:ui} \\
    S &= I_p(X_1,X_2;Y)-\min_{q\in \Delta_p} I_q(X_1,X_2;Y), \label{eqn:si}
\end{align}
where $\Delta_p:=\{ q\in \Delta: q(x_i,y)=p(x_i,y), ~\forall y\in\mathcal{Y}, x_i\in\mathcal{X}_i, i\in[2] \}$ characterizes the set of \textit{marginal-matching} joint distributions, and $I_q$ is the mutual information (MI) over these joint distribution $q(x_1,x_2,y)$. The key lies in optimizing $q \in \Delta_p$ such that the marginals $q(x_1, y) = p(x_1, y)$ and $q(x_2, y) = p(x_2, y)$ are preserved, while relaxing the coupling between $x_1$ and $x_2$; that is, $q(x_1, x_2)$ is not necessarily equal to $p(x_1, x_2)$. The intuition behind this is that redundancy and uniqueness should be identifiable given access only to $p(x_1, y)$ and $p(x_2, y)$, and thus should depend solely on $q \in \Delta_p$. In contrast, synergy inherently depends on the joint distribution $p(x_1, x_2)$, which is reflected in Eq.\ref{eqn:si} relying on the full distribution $p$.

\paragraph{Multimodal Fusion.}
Early approaches to multimodal fusion employed kernel-based methods \citep{bucak2013multiple, chen2014emotion, poria2015deep}, graphical models \citep{nefian2002coupled, garg2003boosted, reiter2007hidden}, and neural networks \citep{ngiam2011multimodal, gao2015talking, nojavanasghari2016deep}. With advances in deep learning, more sophisticated fusion strategies have emerged. Tensor-based methods \citep{zadeh2017tensor, liu2018efficient} perform outer-product fusion to capture multiplicative interactions, while attention-based approaches \citep{rahman2020integrating, yang2021leverage} generate cross-modal displacement vectors through self-attention mechanisms. The Multimodal Transformer (MulT) \citep{tsai2019multimodal} introduced cross-modal attention blocks for word-level alignment across vision, language, and audio. In clinical settings, late fusion approaches \citep{khadanga2019using, deznabi2021predicting} concatenate embeddings from pre-trained encoders, while \citet{soenksen2022integrated} developed a generalizable pipeline for electronic health records spanning four modalities through direct concatenation followed by gradient boosting. More recently, MISTS \citep{zhang2023improving} extended cross-modal attention with multi-time attention modules to handle temporal irregularities.

\begin{table}[h]
\centering
\caption{Comparison between Multimodal MoE and MERGE for massively multimodal learning.}
\label{tab:moe_comparison}
\footnotesize
\setlength{\tabcolsep}{4pt}
\begin{tabular}{@{}lcc@{}}
\toprule
\textbf{Aspect} & \textbf{Multimodal MoE} & \textbf{MERGE} \\
\midrule
Routing Basis & Expert-token similarity & Temporal RUS + Similarity \\
Modality Scalability & $O(M^2)$ implicit & Sub-linear (RUS-guided) \\
Cross-Modal Awareness & Implicit & Explicit pairwise temporal interaction \\
Expert Specialization & Emergent from training & Principled (R/U/S-specific) \\
Temporal Dynamics & Single time point & Multi-lag temporal interaction \\
Routing Decision & Black-box & Interpretable RUS values \\
Computational Efficiency & Standard sparsity & RUS-optimized sparsity \\
Fusion Strategy & Fixed & Adaptive mixture of fusion types \\
\bottomrule
\end{tabular}
\end{table}

\paragraph{MoE-Based Multimodal Fusion.}
MoE architectures have emerged as a promising paradigm for multimodal fusion due to their ability to scale efficiently while enabling expert specialization. LIMoE \citep{mustafa2022multimodal} pioneered large-scale multimodal MoE by processing both images and text through a shared sparse transformer with contrastive learning, demonstrating that experts naturally specialize in different modalities through entropy-based regularization. FuseMoE \citep{han2024fusemoe} addressed the ``FlexiModal'' setting with irregularly sampled and missing modalities, introducing a Laplace gating function with theoretical convergence guarantees superior to softmax routing. Building on this, Flex-MoE \citep{yun2024flex} proposed a missing modality bank and dual-router design ($\mathcal{G}$-Router for generalized knowledge, $\mathcal{S}$-Router for modality-specific combinations) to handle arbitrary modality availability. Hierarchical MoE \citep{nguyen2024expert} demonstrated that Laplace gating at two hierarchical levels eliminates undesirable parameter interactions, accelerating expert convergence in multimodal tasks. I$^2$MoE \citep{xin2025i2moe} incorporated interaction-type awareness by assigning separate experts to redundancy, uniqueness, and synergy interactions using weakly supervised losses derived from Partial Information Decomposition. ConfSMoE \citep{zheng2025rethinking} addressed expert collapse through confidence-guided gating that detaches routing scores from task confidence, providing theoretical insights into why experts fail to specialize under standard softmax routing. Despite these advances, existing multimodal MoE approaches face limitations in \textit{massively multimodal} settings involving dozens to hundreds of heterogeneous input streams. We summarize these challenges and compare key aspects of multimodal MoE with MERGE in Table \ref{tab:moe_comparison}.

% \subsection{Transfer Entropy Estimation}

% \begin{algorithm}
% \caption{Transfer Entropy Estimation}
% \begin{algorithmic}[1]
% \REQUIRE Source time series $X$, target time series $Y$, lag $\tau$, number of bins $B$
% \ENSURE Transfer entropy $TE$
% \STATE Create lagged variables: $X_{past} = X[:-\tau]$, $Y_{past} = Y[:-\tau]$, $Y_{present} = Y[\tau:]$
% \IF{data is continuous}
%     \STATE Discretize using $B$ bins with equal-width binning
% \ENDIF
% \STATE Compute joint distribution $P(Y_{present}, X_{past}, Y_{past})$
% \STATE Compute marginal distribution $P(Y_{present}, Y_{past})$
% \STATE Calculate conditional distributions
% \STATE Compute KL divergence using relative entropy
% \STATE $TE = \sum P(y_t, x_{t-\tau}, y_{t-1}) \log \frac{P(y_t | x_{t-\tau}, y_{t-1})}{P(y_t | y_{t-1})}$
% \RETURN $TE$
% \end{algorithmic}
% \end{algorithm}

% \subsection{Time-Specific Temporal PID Decomposition}

\section{Dataset Information and Processing Details} \label{sec:dataset_supp}
\subsection{MIMIC-IV}
The Medical Information Mart for Intensive Care (MIMIC-IV) ecosystem comprises deidentified clinical databases of patients admitted to the emergency department or intensive care unit (ICU) at Beth Israel Deaconess Medical Center. We use lab measurements, vital signs, and radiology notes from MIMIC-IV and MIMIC-IV-Notes \citep{johnson2023mimic}, along with chest X-rays from MIMIC-CXR \citep{johnson2019mimic,johnson2019mimiccxrjpg}, to construct temporal sequences of ICU stays. Input signals are categorized into three modalities: labs/vitals, clinical notes, and chest X-rays. We extract embeddings from clinical notes using BioBERT \citep{lee2020biobert} and from chest X-rays using a DenseNet \citep{huang2017densely} pre-trained on CheXpert \citep{irvin2019chexpert}. We consider two prediction tasks: in-hospital mortality (IHM) and length-of-stay (LOS). For IHM, a stay is labeled as mortality if the patient died after the first 48 hours, otherwise as survival. For LOS, the goal is to predict whether the patient will leave the ICU alive in less than 96 hours. The train, validation, and test splits consist of 32,435, 6,950, and 6,952 stays, respectively.

\subsection{CMU-MOSI}
CMU-MOSI \citep{zadeh2016mosi} is a multimodal dataset consisting of 2,199 opinion video segments, including speech, visual gestures, and audio, drawn from 93 YouTube vlog videos and annotated for sentiment intensity. We use the preprocessed version from MultiBench \citep{liang2021multibench} and focus on the binary sentiment classification task (positive vs. negative). The modalities considered are vision, text, and audio. The train, validation, and test splits contain 1,283, 214, and 686 segments, respectively.

\subsection{WESAD}
WESAD \citep{schmidt2018introducing} is a multimodal dataset for wearable stress and affect detection, collected from 15 participants in a controlled laboratory study. It includes physiological and motion data from chest- and wrist-worn devices, covering signals such as electrodermal activity (EDA), electrocardiogram (ECG), acceleration, temperature, blood volume pulse (BVP), and respiration. The task is affect recognition across six states: transient, baseline, stress, amusement, meditation, and unknown. We group the signals into two modalities: chest signals and wrist signals. The dataset is split by subject, with 10, 2, and 3 subjects allocated to the train, validation, and test sets, respectively.

\subsection{Opportunity}
The Opportunity dataset \citep{chavarriaga2013opportunity} comprises a comprehensive human activity recognition corpus collected in a sensor-rich kitchen environment, featuring four subjects performing activities of daily living (ADL). The dataset contains 20 experimental sessions, with each subject completing 5 ADL runs (S1-ADL1 through S4-ADL5), while drill sessions were excluded from our experimental protocol. Data collection was conducted at a 30 Hz sampling frequency, yielding 250 feature columns encompassing 243 sensor measurements and 7
  activity labels. The activity recognition task focuses on 5 high-level activity classes: Relaxing (101), Coffee time (102), Early
  morning (103), Cleanup (104), and Sandwich time (105). The preprocessing pipeline implemented a subject-based train/validation/test split protocol to prevent temporal data
  leakage, allocating subjects 1-3 (runs ADL1-ADL4) for training (342,535 samples), subjects 1-3 (run ADL5) for validation (76,399
  samples), and subject 4 (all runs) for testing (111,644 samples). The 243 sensor features were systematically categorized into seven distinct modalities based on
  anatomical location and sensor type: Torso (19 sensors), Arms (58 sensors), Legs (52 sensors), Shoes (32 sensors), Objects (35
  sensors), Environment (34 sensors), and Location (12 sensors). 

\subsection{PAMAP2}
The PAMAP2 Physical Activity Monitoring dataset \citep{reiss2012introducing} comprises a comprehensive multimodal sensor corpus collected from nine participants
  performing 18 distinct physical activities in naturalistic settings. The dataset incorporates heterogeneous sensor modalities
  including three Colibri wireless Inertial Measurement Units (IMUs) positioned at anatomically strategic locations: dominant wrist,
  chest, and dominant ankle, sampling at 100 Hz, complemented by a heart rate monitor operating at approximately 9 Hz. Each
  experimental session captures 54-dimensional feature vectors encompassing temporal timestamps, activity labels, and 52 attributes of
  raw sensory measurements spanning accelerometry (both 16g and 6g ranges), gyroscopy, magnetometry, and quaternion orientation data.
  The activity recognition framework targets 18 physical activities including fundamental locomotion patterns (lying, sitting,
  standing, walking, running), complex motor tasks (cycling, stair navigation), daily living activities (computer work, household
  tasks), and recreational pursuits (soccer, rope jumping).
  Similar to the Opportunity dataset, we implemented a subject-based partitioning strategy to ensure temporal independence and prevent data
  leakage, allocating subjects 1-6 for training, subject 7 for validation, and subjects 8-9 for testing. The 52-dimensional sensor feature space was systematically organized into four modalities: chest, hand, ankle, and heart rate.

\section{Detailed Procedure of Multi-Scale BATCH Estimator} \label{sec:batch_supp}
\begin{algorithm}[H]
\caption{Phase 1: Multi-Task Discriminator Training}
\label{alg:mtl-discriminator}
\begin{algorithmic}[1]
\REQUIRE Multi-lag datasets $\{\mathcal{D}_\tau\}_{\tau=0}^{K-1}$, lag weights $\{w_\tau\}$

\STATE Initialize shared backbone networks and lag-specific components
\FOR{$\text{epoch} = 1$ to $N_{\text{disc}}$}
    \FOR{$\text{step} = 1$ to $\text{steps\_per\_epoch}$}
        \STATE Sample lag $\tau \sim \text{Categorical}(\{w_\tau\})$
        \STATE Sample batch $(x_{1,\tau}^{(b)}, x_{2,\tau}^{(b)}, y^{(b)}) \sim \mathcal{D}_\tau$

        \STATE \textbf{Forward pass with lag conditioning:}
        \STATE $\text{logits}_1 \leftarrow D_{1,\theta}(x_{1,\tau}^{(b)}, \tau)$
        \STATE $\text{logits}_2 \leftarrow D_{2,\theta}(x_{2,\tau}^{(b)}, \tau)$
        \STATE $\text{logits}_{12} \leftarrow D_{12,\theta}([x_{1,\tau}^{(b)}; x_{2,\tau}^{(b)}], \tau)$

        \STATE \textbf{Compute discriminator loss:}
        \STATE $\mathcal{L}_{\text{batch}} \leftarrow H(y^{(b)}, \text{logits}_1) + H(y^{(b)}, \text{logits}_2) + H(y^{(b)}, \text{logits}_{12})$

        \STATE Update $\theta$ via $\nabla_\theta \mathcal{L}_{\text{batch}}$
    \ENDFOR
\ENDFOR

\RETURN Trained discriminator parameters $\theta_{\text{disc}}$
\end{algorithmic}
\end{algorithm}

\begin{algorithm}[H]
\caption{Phase 2: Multi-Task Alignment for $Q$}
\label{alg:mtl-alignment}
\begin{algorithmic}[1]
\REQUIRE Multi-lag datasets $\{\mathcal{D}_\tau\}_{\tau=0}^{K-1}$, lag weights $\{w_\tau\}$, trained discriminators $\theta_{\text{disc}}$

\STATE Initialize lag-conditioned Alignment module with parameters $\theta_{\text{align}}$
\STATE Freeze discriminator parameters $\theta_{\text{disc}}$
\FOR{$\text{epoch} = 1$ to $N_{\text{align}}$}
    \FOR{$\text{step} = 1$ to $\text{steps\_per\_epoch}$}
        \STATE Sample lag $\tau \sim \text{Categorical}(\{w_\tau\})$
        \STATE Sample batch $(x_{1,\tau}^{(b)}, x_{2,\tau}^{(b)}, y^{(b)}) \sim \mathcal{D}_\tau$

        \STATE \textbf{Compute lag-conditioned embeddings:}
        \STATE $q_{X_1} \leftarrow \phi_1(g_{1,\theta}(x_{1,\tau}^{(b)}), e(\tau))$
        \STATE $q_{X_2} \leftarrow \phi_2(g_{2,\theta}(x_{2,\tau}^{(b)}), e(\tau))$
        
        \STATE \textbf{Compute lag-specific alignment:}
        \STATE $\text{align}_\tau[i,j,k] \leftarrow \exp\left(\frac{q_{X_1}^{(i,k)} \cdot q_{X_2}^{(j,k)}}{\sqrt{d}}\right)$
        \STATE \textbf{Sinkhorn normalization with lag-specific discriminators:}
        \STATE $P_{Y|X_1,\tau} \leftarrow \text{Softmax}(D_{1,\theta_{\text{disc}}}(x_1, \tau))$
        \STATE $P_{Y|X_2,\tau} \leftarrow \text{Softmax}(D_{2,\theta_{\text{disc}}}(x_2, \tau))$
        
        \FOR{$k = 1$ to $C$}
            \FOR{$\text{iter} = 1$ to $\text{max\_iter}$}
                \STATE $Q_\tau \leftarrow$ Apply Sinkhorn updates on $\text{align}_\tau[:,:,k]$ using $P_{Y|X_1,\tau}$ and $P_{Y|X_2,\tau}$
                \STATE Check convergence
            \ENDFOR
        \ENDFOR

        \STATE \textbf{Compute alignment loss:}
        \STATE $\mathcal{L}_{\text{align}} \leftarrow \text{AlignmentLoss}(Q_\tau, x_{1,\tau}^{(b)}, x_{2,\tau}^{(b)}, y^{(b)})$

        \STATE Update alignment parameters $\theta_{\text{align}}$ via $\nabla_{\theta_{\text{align}}} \mathcal{L}_{\text{align}}$
    \ENDFOR
\ENDFOR

\RETURN Trained alignment parameters $\theta_{\text{align}}$
\end{algorithmic}
\end{algorithm}

\begin{algorithm}[H]
\caption{Phase 3: Multi-Lag RUS Computation}
\label{alg:mtl-rus-computation}
\begin{algorithmic}[1]
\REQUIRE Multi-lag datasets $\{\mathcal{D}_\tau\}_{\tau=0}^{K-1}$, trained discriminators $\theta_{\text{disc}}$, trained alignment module $\theta_{\text{align}}$

\FOR{$\tau \in \{0, 1, \ldots, K-1\}$}
    \STATE \textbf{Compute mutual information (MI) terms using trained discriminators:}
    \STATE $\text{MI}_\tau(Y; X_1) \leftarrow \text{EstimateMI}(D_{1,\theta_{\text{disc}}}(\cdot, \tau), \mathcal{D}_\tau)$
    \STATE $\text{MI}_\tau(Y; X_2) \leftarrow \text{EstimateMI}(D_{2,\theta_{\text{disc}}}(\cdot, \tau), \mathcal{D}_\tau)$
    \STATE $\text{MI}_\tau(Y; X_1, X_2) \leftarrow \text{EstimateMI}(D_{12,\theta_{\text{disc}}}(\cdot, \tau), \mathcal{D}_\tau)$

    \STATE \textbf{Compute optimal alignment distribution:}
    \STATE $Q_\tau^* \leftarrow \text{ComputeOptimalAlignment}(\theta_{\text{align}}, \mathcal{D}_\tau, \tau)$
    \STATE $\text{MI}_{Q_\tau^*}(Y; X_1, X_2) \leftarrow \text{EstimateMI}(Q_\tau^*, \mathcal{D}_\tau)$

    \STATE \textbf{Decompose into RUS components:}
    \STATE $R_\tau \leftarrow \text{MI}_\tau(Y; X_1) + \text{MI}_\tau(Y; X_2) - \text{MI}_{Q_\tau^*}(Y; X_1, X_2)$ \COMMENT{Redundancy}
    \STATE $U_{1,\tau} \leftarrow \text{MI}_{Q_\tau^*}(Y; X_1, X_2) - \text{MI}_\tau(Y; X_2)$ \COMMENT{Uniqueness $X_1$}
    \STATE $U_{2,\tau} \leftarrow \text{MI}_{Q_\tau^*}(Y; X_1, X_2) - \text{MI}_\tau(Y; X_1)$ \COMMENT{Uniqueness $X_2$}
    \STATE $S_\tau \leftarrow \text{MI}_\tau(Y; X_1, X_2) - \text{MI}_{Q_\tau^*}(Y; X_1, X_2)$ \COMMENT{Synergy}

\ENDFOR

\RETURN Temporal RUS components $\{R_\tau, U_{1,\tau}, U_{2,\tau}, S_\tau\}_{\tau=0}^{K-1}$
\end{algorithmic}
\end{algorithm}

\section{Model Architecture Details} \label{sec:architecture_supp}
\subsection{Modality-Specific Encoder}
\begin{table}[H]
\centering
\begin{tabular}{lp{8cm}}
\toprule
\textbf{Component} & \textbf{Configuration} \\
\midrule
Input Projection Layer & Linear transformation from modality dimension $D_m$ to model dimension $d_{model}$ \\
Feature Scaling & Input embeddings scaled by $\sqrt{d_{model}}$ for training stability \\
Temporal Convolution (Optional) & Two 1D convolutional layers with residual connections for temporal feature extraction \\
Convolutional Architecture & 1D convolution: $d_{model} \rightarrow d_{model}$ with kernel size $k$ \\
Batch Normalization & Applied after each convolutional layer \\
Activation Function & ReLU activation for non-linearity \\
Self-Attention Layers & Multi-layer transformer encoder \\
Output Normalization & Layer normalization applied to final embeddings \\
\bottomrule
\end{tabular}
\caption{Modality-Specific Encoder Architecture Components}
\end{table}

\begin{table}[H]
\centering
\begin{tabular}{lcl}
\toprule
\textbf{Parameter} & \textbf{Value} & \textbf{Description} \\
\midrule
Encoder Layers ($L_{enc}$) & 2 & Number of transformer encoder layers \\
Attention Heads ($H$) & 4 & Multi-head attention heads per layer \\
Feed-Forward Dimension ($d_{ff}$) & 256 & Inner dimension of position-wise FFN \\
Dropout Rate ($p_{drop}$) & 0.1 & Dropout probability for regularization \\
Convolution Kernel ($k$) & 3 & Temporal convolution kernel size \\
\bottomrule
\end{tabular}
\caption{Modality-Specific Encoder Hyperparameters}
\end{table}

\subsection{Expert Networks}
\begin{table}[H]
\centering
\begin{tabular}{lp{8cm}}
\toprule
\textbf{Layer} & \textbf{Transformation} \\
\midrule
Input Layer & Linear transformation: $\mathbb{R}^{d_{model}} \rightarrow \mathbb{R}^{d_{expert}}$ \\
Activation & ReLU non-linear activation function \\
Regularization & Dropout with probability $p_{drop}$ \\
Output Layer & Linear transformation: $\mathbb{R}^{d_{expert}} \rightarrow \mathbb{R}^{d_{model}}$ \\
\bottomrule
\end{tabular}
\caption{Feed-Forward Expert Network Architecture}
\end{table}

\begin{table}[H]
\centering
\begin{tabular}{lp{8cm}}
\toprule
\textbf{Component} & \textbf{Description} \\
\midrule
Multi-Head Self-Attention & Scaled dot-product attention with $H$ parallel heads \\
Attention Residual & Residual connection around attention sub-layer \\
Post-Attention Norm & Layer normalization after attention residual \\
Position-wise FFN & Two-layer feed-forward network with ReLU \\
FFN Residual & Residual connection around feed-forward sub-layer \\
Final Normalization & Layer normalization after FFN residual \\
Dropout Regularization & Applied to attention output and FFN output \\
\bottomrule
\end{tabular}
\caption{Synergy Expert Network Architecture}
\end{table}

\begin{table}[H]
\centering
\begin{tabular}{lcl}
\toprule
\textbf{Parameter} & \textbf{Default} & \textbf{Description} \\
\midrule
Expert Hidden Dimension ($d_{expert}$) & 128 & Internal processing dimension \\
Synergy Expert Heads ($H_{syn}$) & 4 & Attention heads for synergy experts \\
Expert Dropout ($p_{expert}$) & 0.1 & Dropout rate within expert networks \\
Synergy Experts ($N_{syn}$) & 2 & Number of attention-based synergy experts \\
Total Experts ($N_{expert}$) & 8 & Total number of expert networks \\
\bottomrule
\end{tabular}
\caption{Expert Network Hyperparameters}
\end{table}

\subsection{RUS-Aware Router}
\begin{table}[H]
\centering
\begin{tabular}{lp{8cm}}
\toprule
\textbf{Component} & \textbf{Description} \\
\midrule
Token Processing & Linear projection: $\mathbb{R}^{d_{model}} \rightarrow \mathbb{R}^{d_{token}}$ with ReLU activation \\
Query Generation & Projects processed tokens to attention query space: $\mathbb{R}^{d_{token}} \rightarrow \mathbb{R}^{d_k}$ \\
RUS Key Projection & Maps pairwise RUS values to attention keys: $\mathbb{R}^2 \rightarrow \mathbb{R}^{d_k}$ \\
RUS Value Projection & Maps pairwise RUS values to attention values: $\mathbb{R}^2 \rightarrow \mathbb{R}^{d_v}$ \\
Temporal RUS Encoder & Single-layer GRU processes concatenated uniqueness and attention context \\
Routing MLP & Two-layer network outputs expert routing logits \\
\bottomrule
\end{tabular}
\caption{RUS-Aware Routing Network Architecture}
\end{table}

\begin{table}[H]
\centering
\begin{tabular}{lcl}
\toprule
\textbf{Parameter} & \textbf{Value} & \textbf{Description} \\
\midrule
GRU Hidden Dimension ($d_{gru}$) & 64 & Temporal context encoding dimension \\
Token Processing Dimension ($d_{token}$) & 64 & Intermediate token representation size \\
Attention Key Dimension ($d_k$) & 32 & Query-key attention space dimension \\
Attention Value Dimension ($d_v$) & 32 & RUS value projection dimension \\
\bottomrule
\end{tabular}
\caption{Routing Network Hyperparameters}
\end{table}

\section{Choice of Hyper-parameters in Auxiliary Losses}
In this section, we discuss the roles of the hyperparameters $\tau$ and $\lambda$ that appear in the auxiliary loss terms in Eq.\ref{eq:r_loss}–Eq.\ref{eq:s_loss}, and provide guidance on how to tune them for improved performance in multimodal applications. The parameters $(\tau_R, \tau_U, \tau_S)$ serve as threshold values that determine when to activate different routing behaviors based on the RUS scores. Conceptually, they control the selectivity of the routing mechanism. If a threshold is set too low, almost all interactions will trigger specialized routing; if set too high, the routing reverts to a standard MoE regime and effectively ignores RUS-guided specialization. Empirically, synergistic interactions tend to be much rarer than redundancy or uniqueness in real-world multimodal data, so we typically set $\tau_S$ lower to avoid overly suppressing synergy-driven routing and to maintain a balanced distribution of routing patterns.

The parameters $(\lambda_R, \lambda_U, \lambda_S)$ regulate the strength of the auxiliary loss terms corresponding to each interaction type. These can be selected based on domain knowledge, e.g., whether a dataset is dominated by synergistic patterns, highly independent modalities, etc., or adjusted to explicitly encourage certain forms of interaction. They can also be made adaptive during training, with validation performance guiding the adjustment.

Although $\tau$ and $\lambda$ encode different aspects of the system, they are related in effect: increasing a threshold $\tau$ reduces how often a particular interaction pattern is activated, which is analogous to decreasing the corresponding $\lambda$, thereby weakening its influence in training.

In practice, we find that because the empirical distribution of RUS values is highly non-uniform, the thresholds $\tau$ (typically set as specific percentiles of their respective RUS distributions) are relatively insensitive to fine-grained tuning. Therefore, we precompute and fix $\tau$ based on dataset statistics and only tune the $\lambda$ values during model training.

\section{Additional Experiments}
\subsection{Synthetic Temporal RUS}
\begin{figure}[h]
    \centering
    \includegraphics[width=0.8\linewidth]{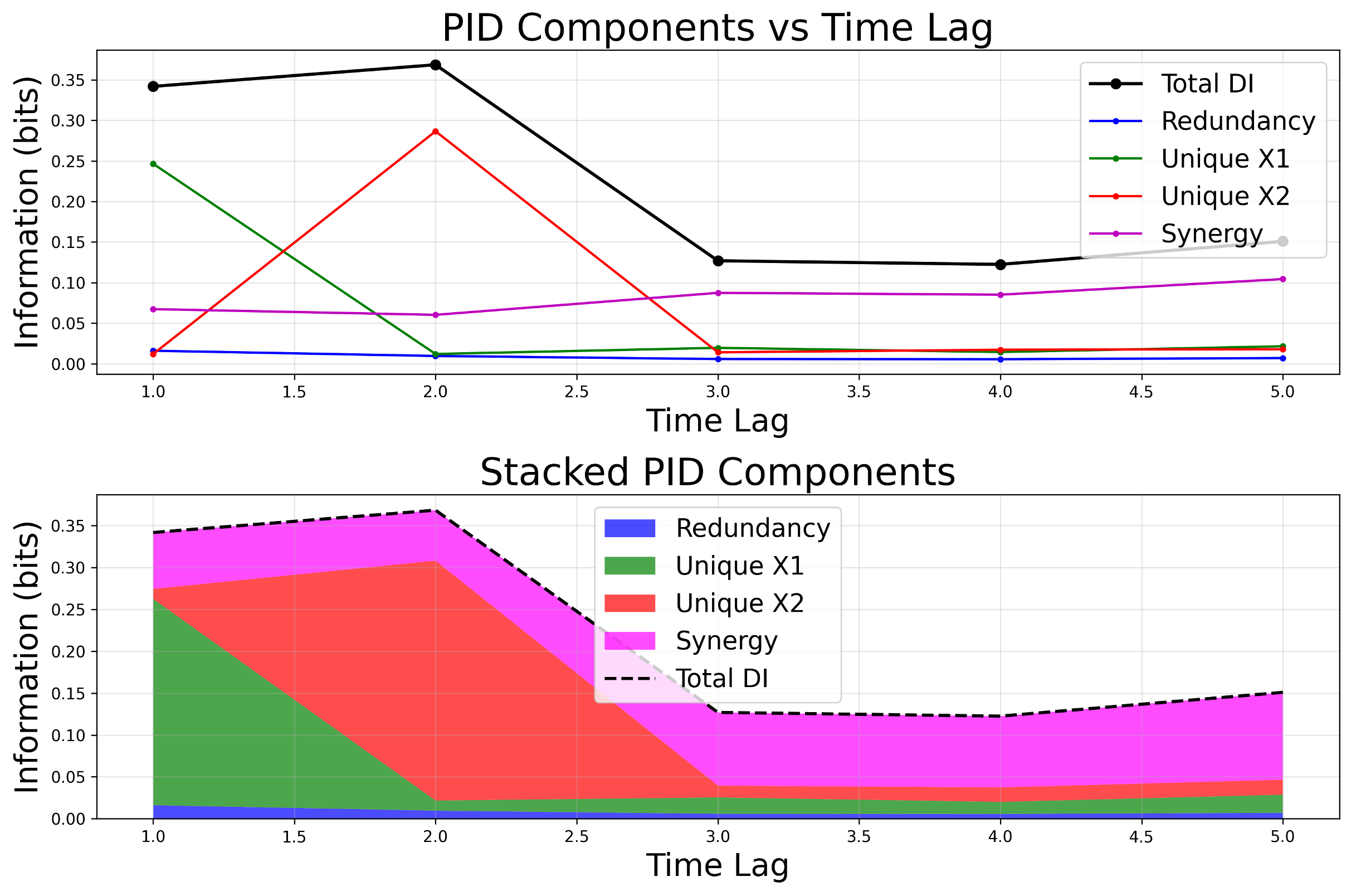}
    \caption{Ground truth temporal RUS of synthetic dataset: at $t = 1$, uniqueness of $X_1$ dominates the total directed information; at $t = 2$, uniqueness of $X_2$ dominates; from $t = 3$ and afterwards, synergy between $X_1$ and $X_2$ dominates the total directed information.}
    \label{fig:trus_synthetic}
\end{figure}

\textbf{Synthetic data generation}~
To validate our multi-scale BATCH estimator framework for computing temporal RUS, we created a synthetic benchmark with known R/U/S values. This synthetic dataset consists of three time series: two independent source variables $X_1$ and $X_2$, and a target variable $Y$. Both $X_1$ and $X_2$ were independently sampled from $N(0, 1)$. $Y$ was constructed with explicit temporal causal influences from both sources, where $X_1$ influences $Y$ with a time lag of 1 ($\tau=1$) and $X_2$ influences $Y$ with a time lag of 2 ($\tau=2$). At each time point $t$, $Y(t)$ was computed as: $Y(t) = \alpha \cdot X_1(t-1) + \alpha \cdot X_2(t-2) + \eta \cdot \epsilon(t)$, where $\epsilon(t)$ is noise sampled from $N(0,1)$. We followed the definition of direction information \citep{weissman2012directed} and information decomposition \citep{bertschinger2014quantifying} to compute temporal RUS values. Figure \ref{fig:trus_synthetic} demonstrates an example of the temporal RUS values we obtained for the synthetic experiments.

\begin{figure}[h]
    \centering
    \includegraphics[width=0.85\linewidth]{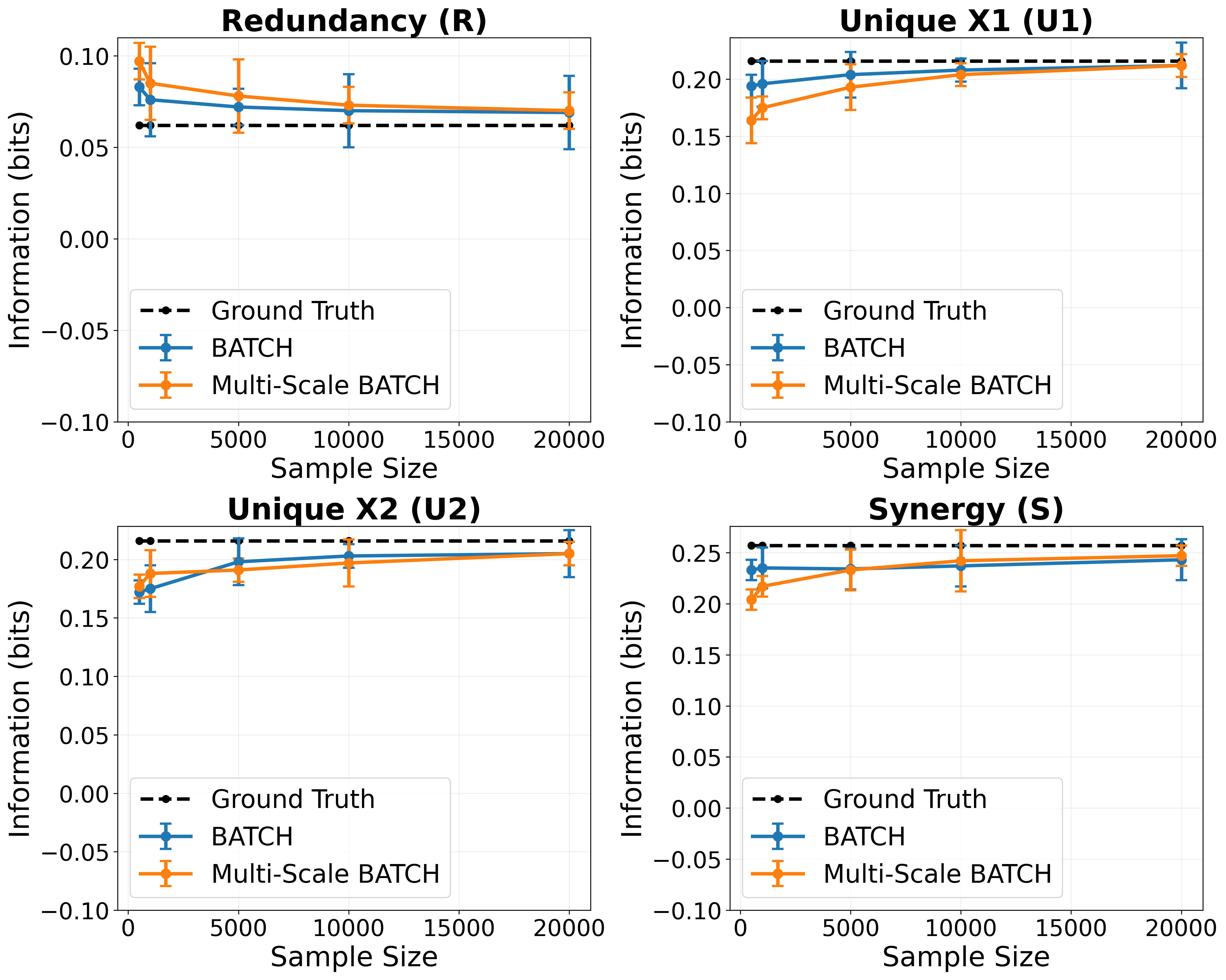}
    \caption{Ablation study of the multi-scale BATCH estimator under different sample sizes: we tested both methods on $[500, 1000, 5000, 10000, 20000]$ samples. Results show that the multi-scale BATCH estimator leads to similar results with step-wise BATCH estimator. As sample size increases, both two methods achieve better approximations to ground truth R/U/S values.}
    \label{fig:sample_r}
\end{figure}

\begin{figure}[h]
    \centering
    \includegraphics[width=0.85\linewidth]{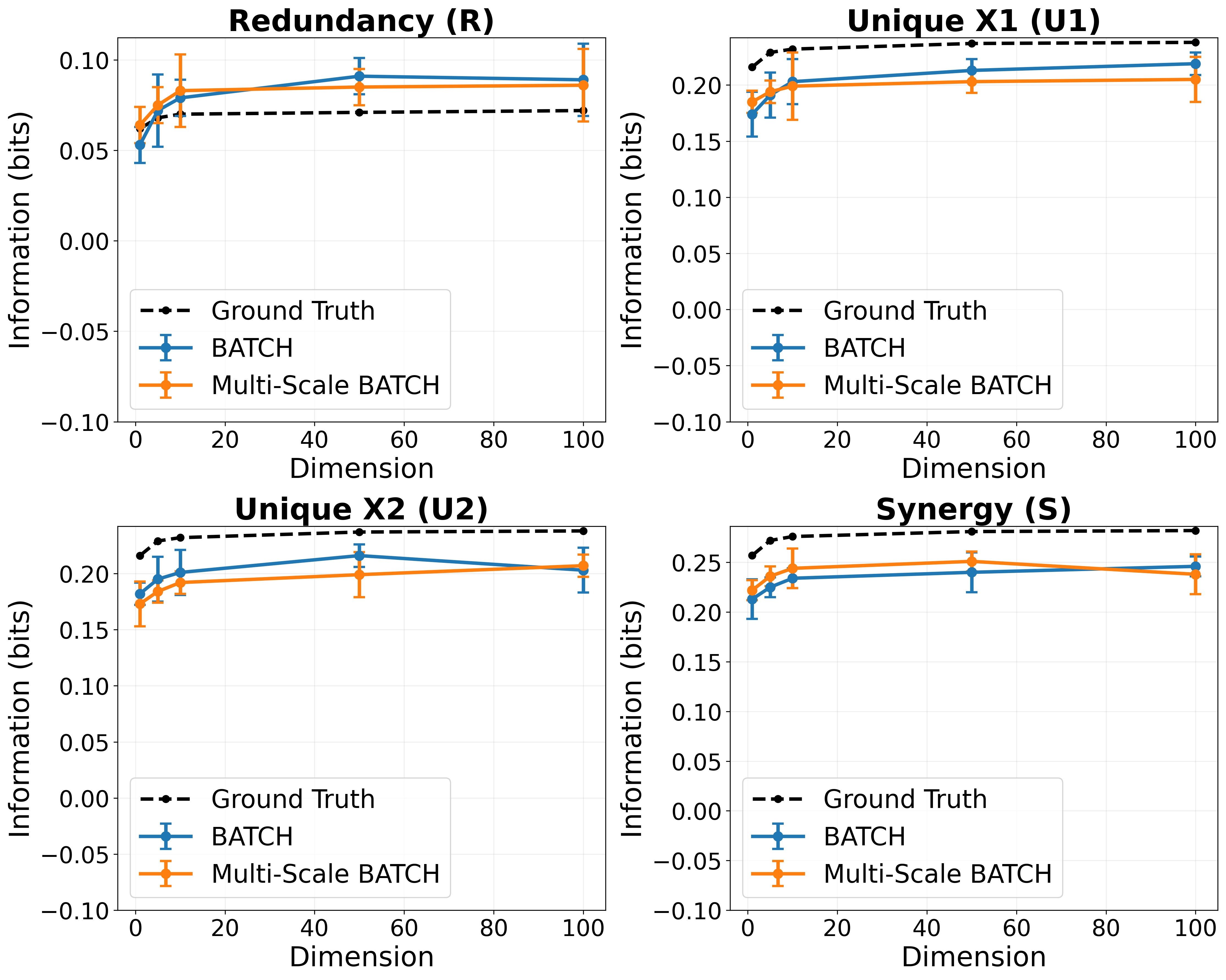}
    \caption{Ablation study on feature dimensionality. We evaluated the multi-scale BATCH estimator under varying feature dimensions $[1, 5, 10, 50, 100]$. Note that the ground-truth R/U/S values naturally change as the dimensionality of the features increases. Across all tested dimensions, both the multi-scale BATCH estimator and the step-wise computation produce consistent R/U/S estimates.}
    \label{fig:dimension_r}
\end{figure}

\begin{figure}[h]
    \centering
    \includegraphics[width=0.85\linewidth]{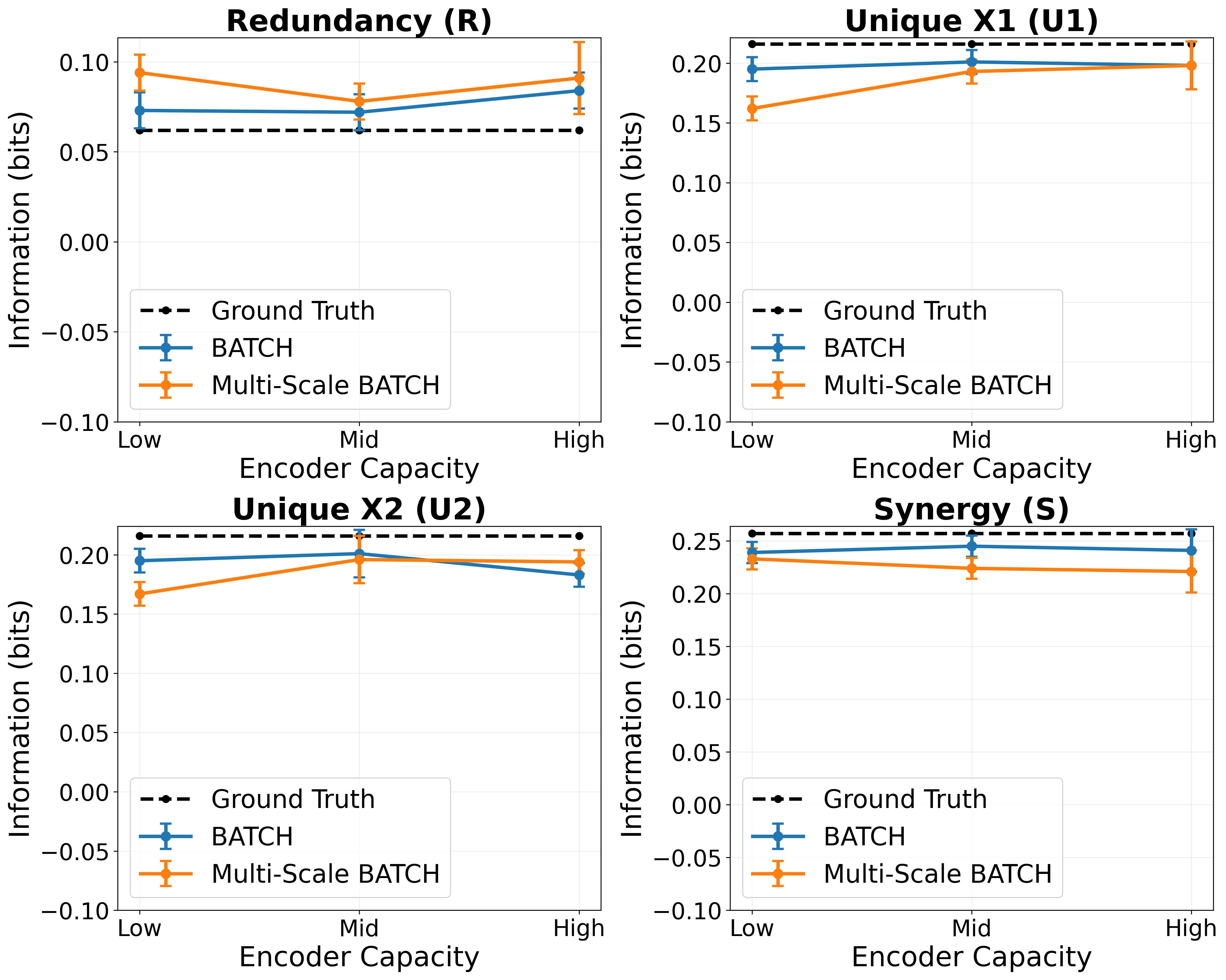}
    \caption{Ablation study on encoder capacity. We further tested the effect of encoder capacity in the BATCH estimator by evaluating 3 configurations: low, medium, and high capacity. In the low-capacity setting, the MLP encoder uses a hidden dimension of 16 with 1 layer; the medium-capacity setting uses a hidden dimension of 32 with 2 layers; and the high-capacity setting uses a hidden dimension of 64 with 3 layers. Across all tested scenarios, we observe that the low- and medium-capacity encoders are generally sufficient for accurate synthetic R/U/S estimation, while increasing the capacity yields only marginal improvements.}
    \label{fig:capacity_r}
\end{figure}

\begin{figure}[h]
    \centering
    \includegraphics[width=0.85\linewidth]{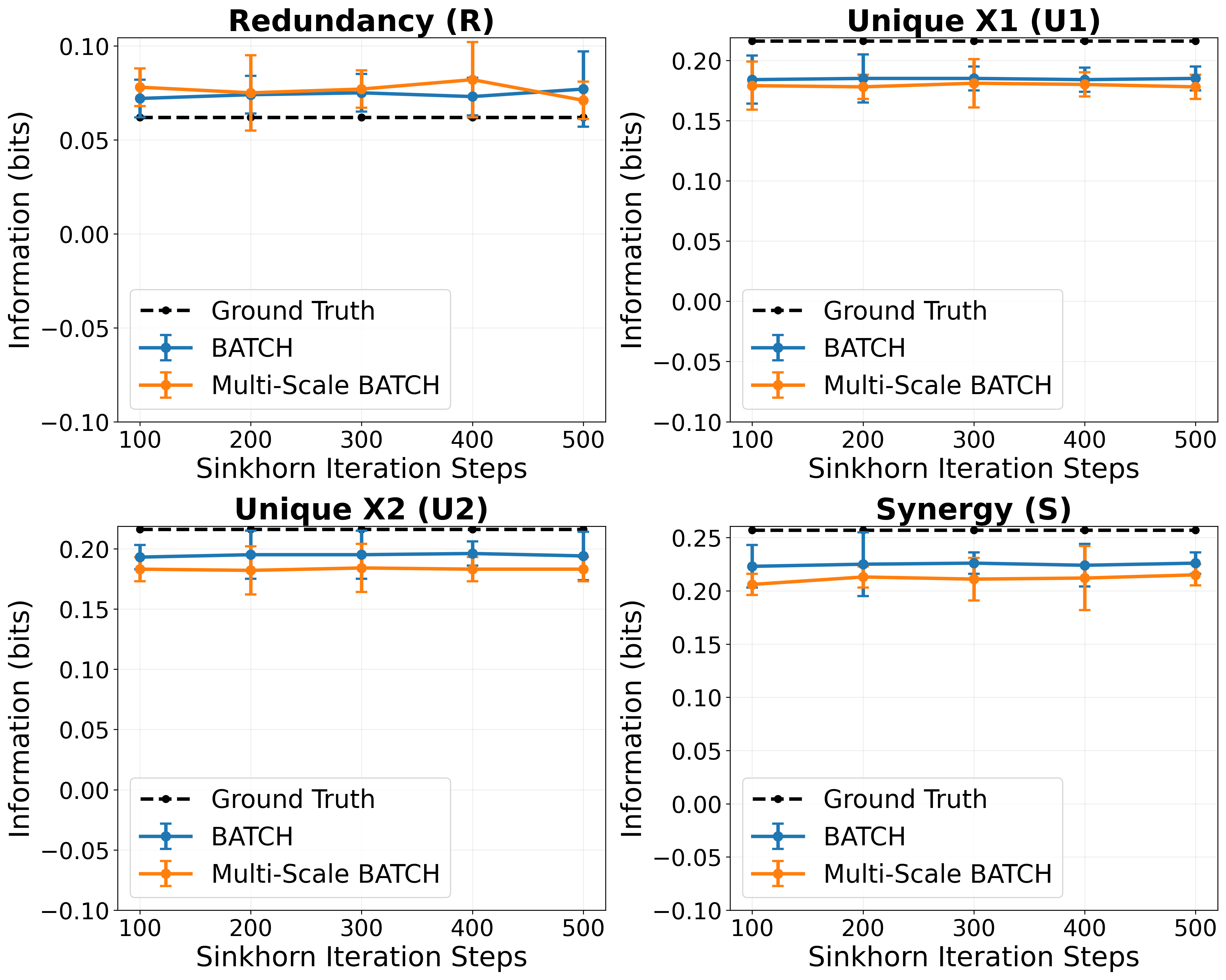}
    \caption{Ablation study on the number of Sinkhorn iteration steps. We evaluated both methods using a range of Sinkhorn iteration counts: $[100, 200, 300, 400, 500]$. Across all settings, the two methods produce close R/U/S estimates, and the estimator seems relatively insensitive to the choice of iteration count.}
    \label{fig:sinkhorn_r}
\end{figure}

We compare the proposed multi-scale BATCH estimator with the step-wise temporal RUS computation using data with known RUS. Our evaluation spans multiple aspects, including varying sample sizes, different feature dimensions for $X_1/X_2$, encoder capacity within the BATCH estimator, and the number of Sinkhorn iterations used during optimization. Figure \ref{fig:sample_r} to Figure \ref{fig:sinkhorn_r} demonstrate these results: all experiments are averaged across 5 random runs with different time lag configurations; the reported R/U/S values are also aggregated across time lags.

\subsection{Effect of Bad R/U/S Estimations}
\begin{figure}[h]
    \begin{minipage}{0.95\textwidth}
    \centering
    \begin{tabular}{@{\hspace{-2.4ex}} c @{\hspace{-0.5ex}} @{\hspace{-2.4ex}} c @{\hspace{-0.5ex}} @{\hspace{-2.4ex}} c @{\hspace{-0.5ex}} @{\hspace{-2.4ex}} c @{\hspace{-2.4ex}}}
        \begin{tabular}{c}
        \includegraphics[width=.27\textwidth]{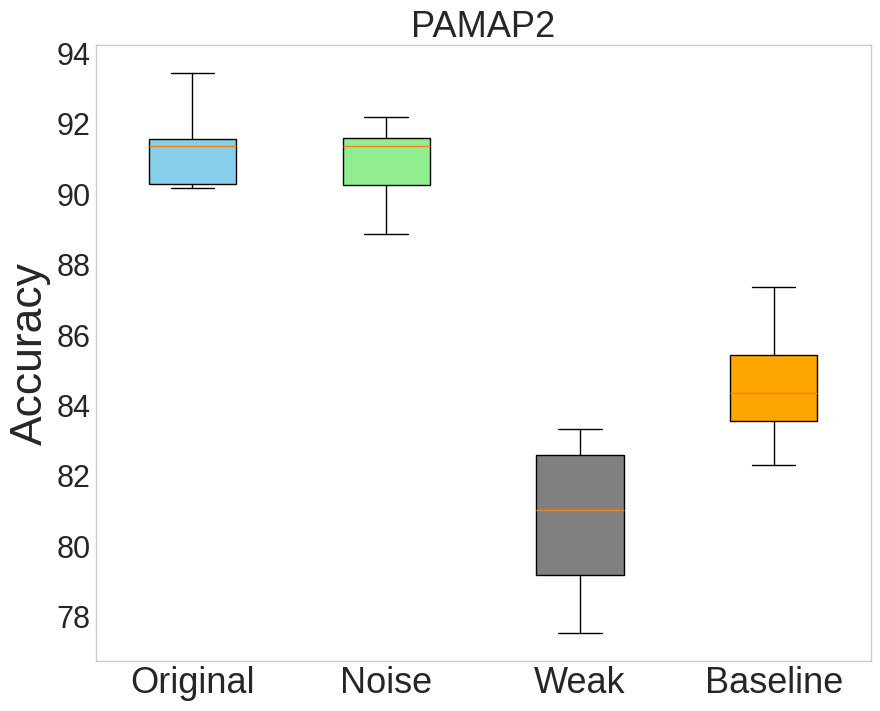}
        %\vspace{-5pt}
        \\
        {\small{(a) PAMAP2}}
        \end{tabular} &
        \begin{tabular}{c}
        \includegraphics[width=.27\textwidth]{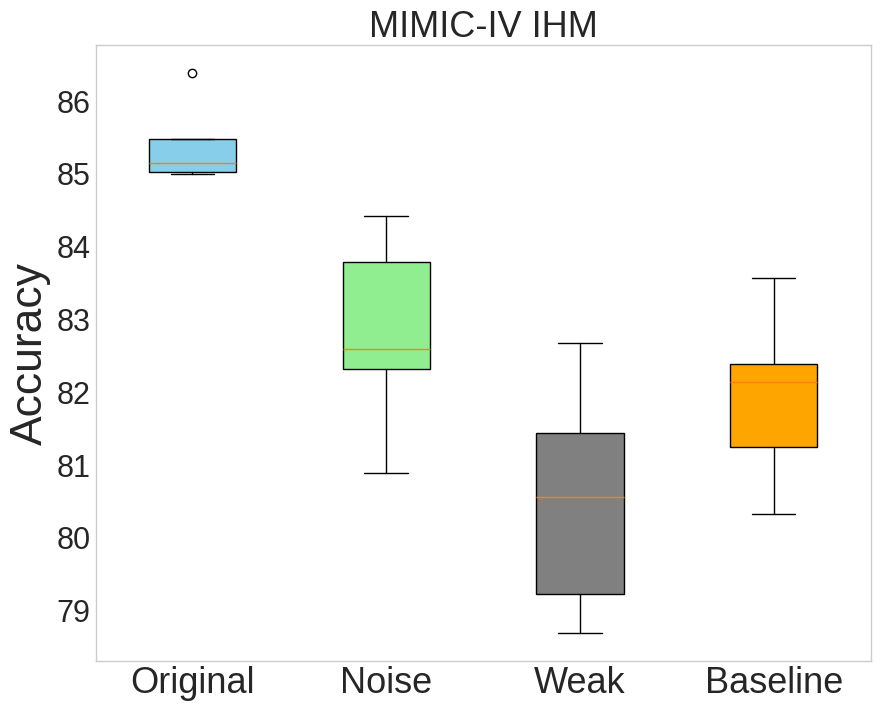}
        %\vspace{-5pt}
        \\
        {\small{(b) MIMIC-IV IHM}}
        \end{tabular} & 
        \begin{tabular}{c}
        \includegraphics[width=.27\textwidth]{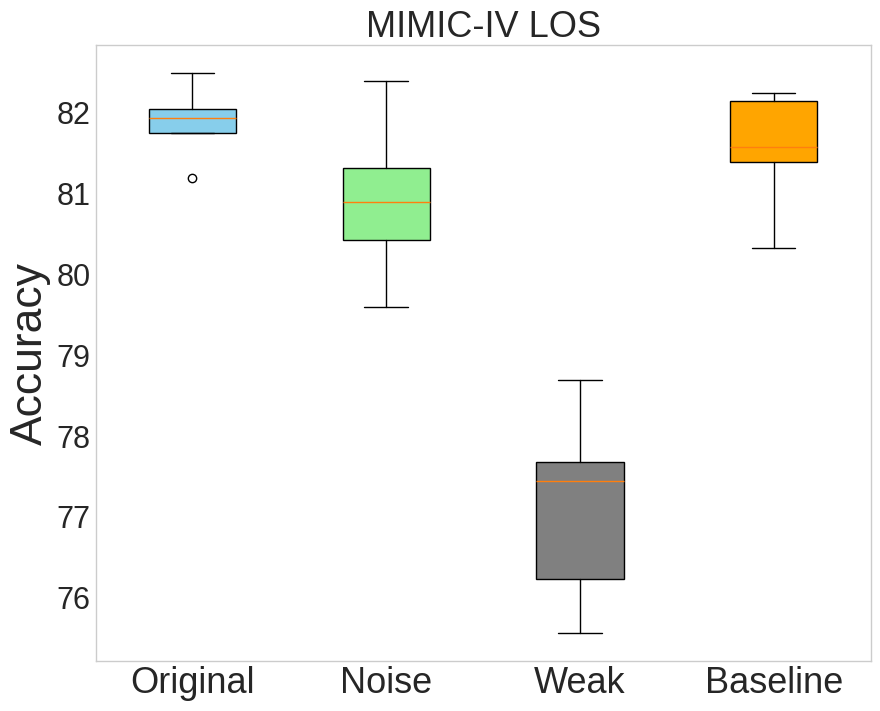} 
        %\vspace{-5pt}
        \\
        {\small{(c) MIMIC-IV LOS}}
        \end{tabular} &
        \begin{tabular}{c}
        \includegraphics[width=.27\textwidth]{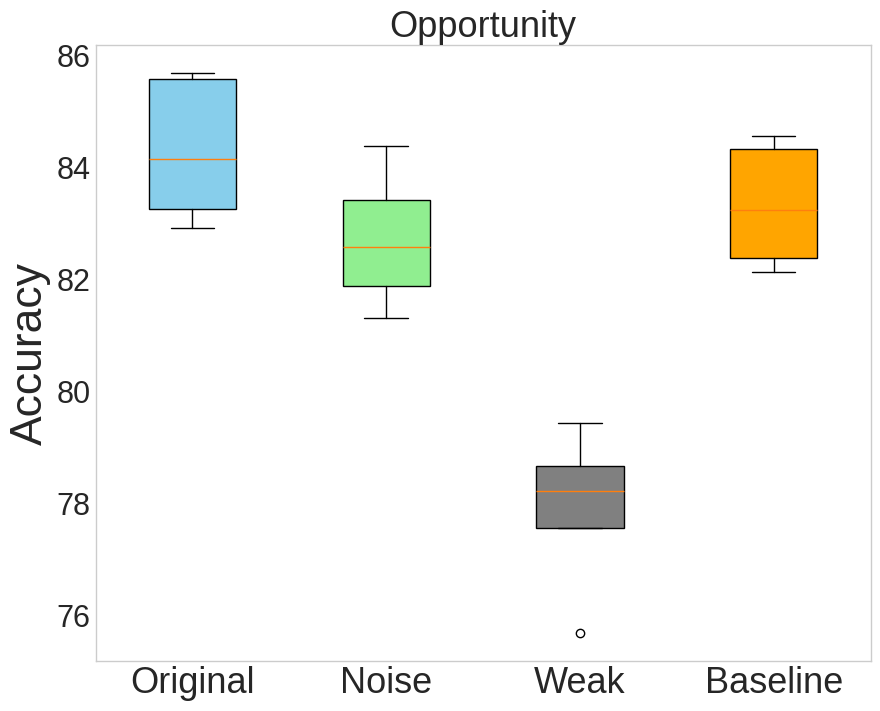}
        %\vspace{-5pt}
        \\
        {\small{(d) Opportunity}}
        \end{tabular} \\
        \end{tabular}
    \end{minipage}
    \vspace{-1em}
    \caption{Effect of inaccurate R/U/S estimation on model performance. Across the four evaluated datasets, we investigated how degrading the quality of the temporal R/U/S estimates impacts \name’s performance. We observed that adding Gaussian noise to the R/U/S sequences has mild impact on downstream performance. However, making the R/U/S sequences weak or sparse leads to substantial performance degradation, often falling below baseline models. We speculate that noisy R/U/S values still preserve the overall structure of modality interactions, enabling the router to make approximately correct decisions. However, sparse or near-zero R/U/S sequences fail to provide meaningful interaction signals. As a result, the model receives misleading guidance and may adopt incorrect or inconsistent routing patterns.}
    \label{fig:bad_rus}
\end{figure}

We conducted additional experiments to assess how sensitive \name~is to inaccuracies in the temporal RUS estimates, particularly in scenarios where the RUS signals are weak or noisy. To this end, we evaluated two variants of the model: (1) Noisy RUS: we injected Gaussian noise into the computed temporal RUS sequences to simulate unstable estimation. (2) Sparse / weakened RUS: we zeroed out a majority of the temporal RUS sequence values, degrading them into sparse interaction patterns. Figure \ref{fig:bad_rus} demonstrates the results.

\subsection{Training Time \& Max GPU Memory}
\begin{figure}[h]
    \begin{minipage}{\textwidth}
    \centering
    % \vspace{-2ex}
    \begin{tabular}{@{\hspace{-3.8ex}} c @{\hspace{-2.4ex}} c @{\hspace{-1.5ex}} c @{\hspace{-1.5ex}}}
        \begin{tabular}{c}
        \includegraphics[width=.31\textwidth]{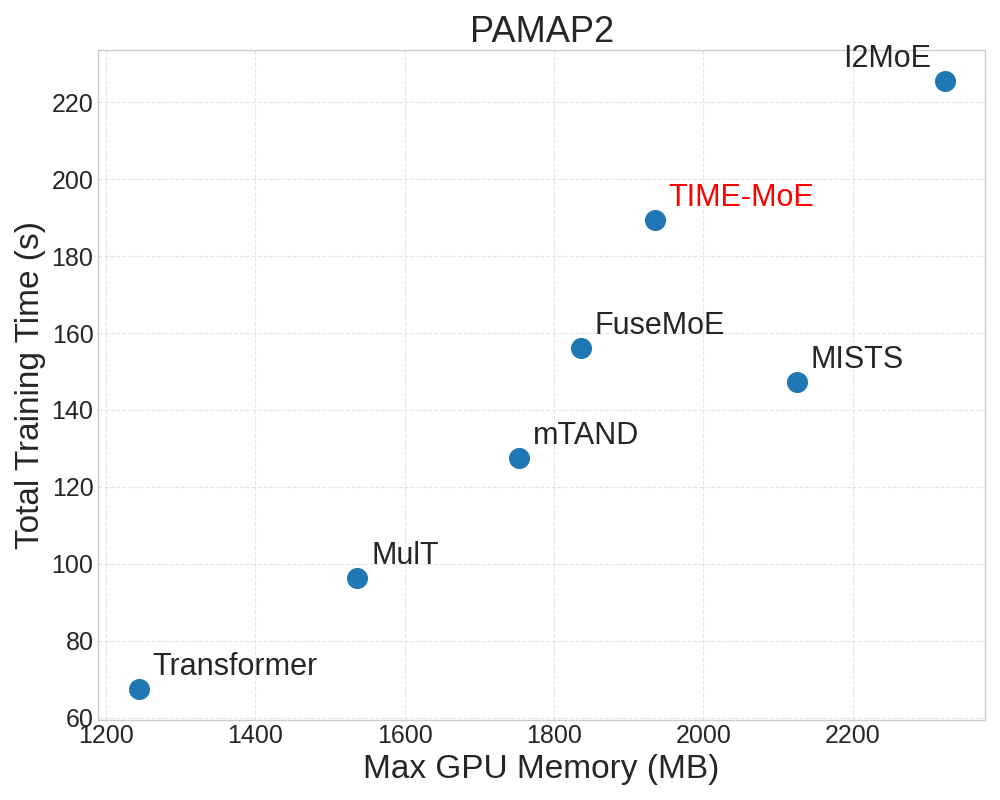}
        %\vspace{-5pt}
        \\
        {\small{(a)}}
        \end{tabular} & 
        \begin{tabular}{c}
        \includegraphics[width=.31\textwidth]{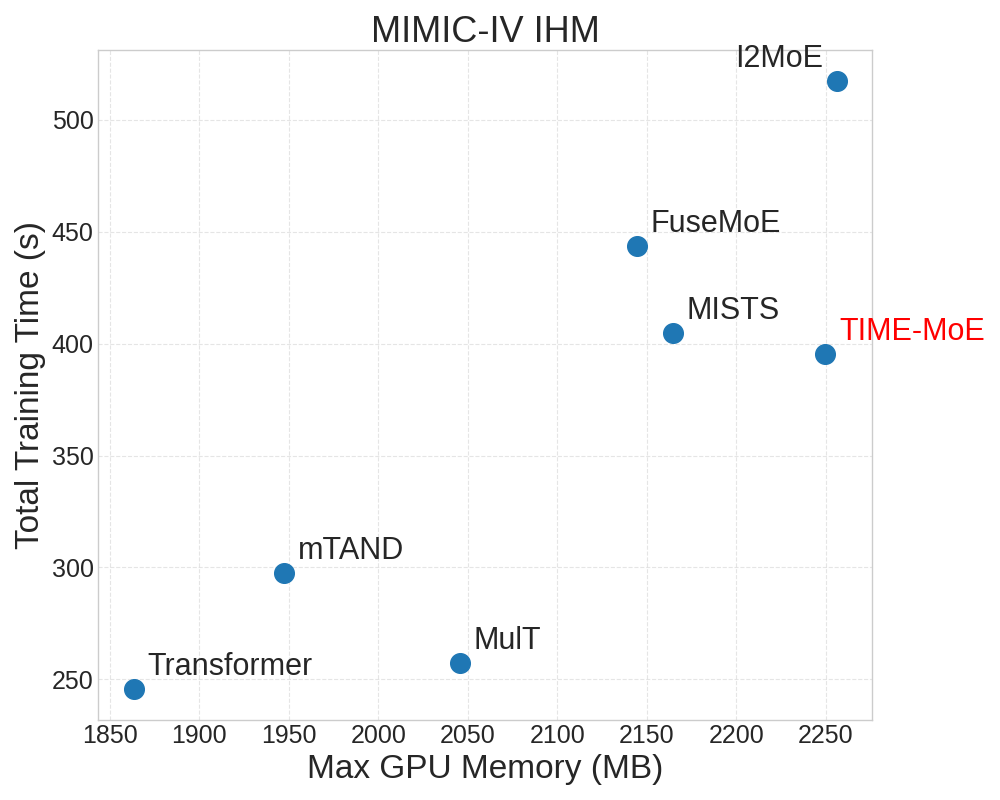} 
        %\vspace{-5pt}
        \\
        {\small{(b)}}
        \end{tabular} &
        \begin{tabular}{c}
        \includegraphics[width=.31\textwidth]{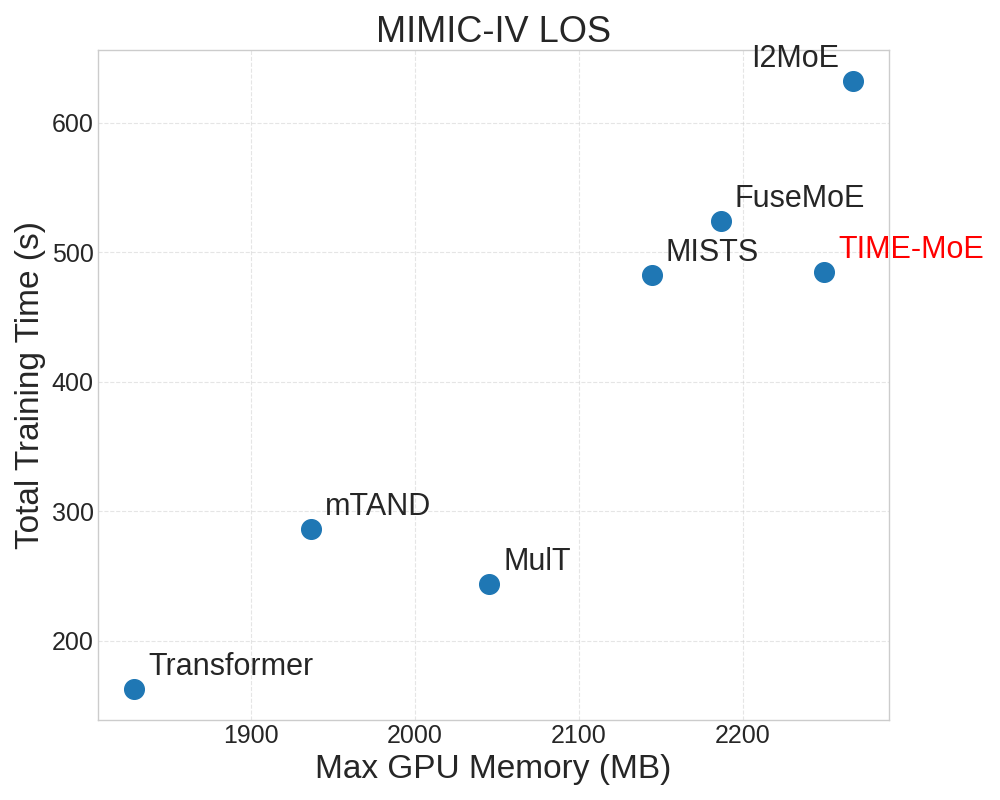} 
        %\vspace{-5pt}
        \\
        {\small{(c)}}
        \end{tabular} \\
        \begin{tabular}{c}
        \includegraphics[width=.31\textwidth]{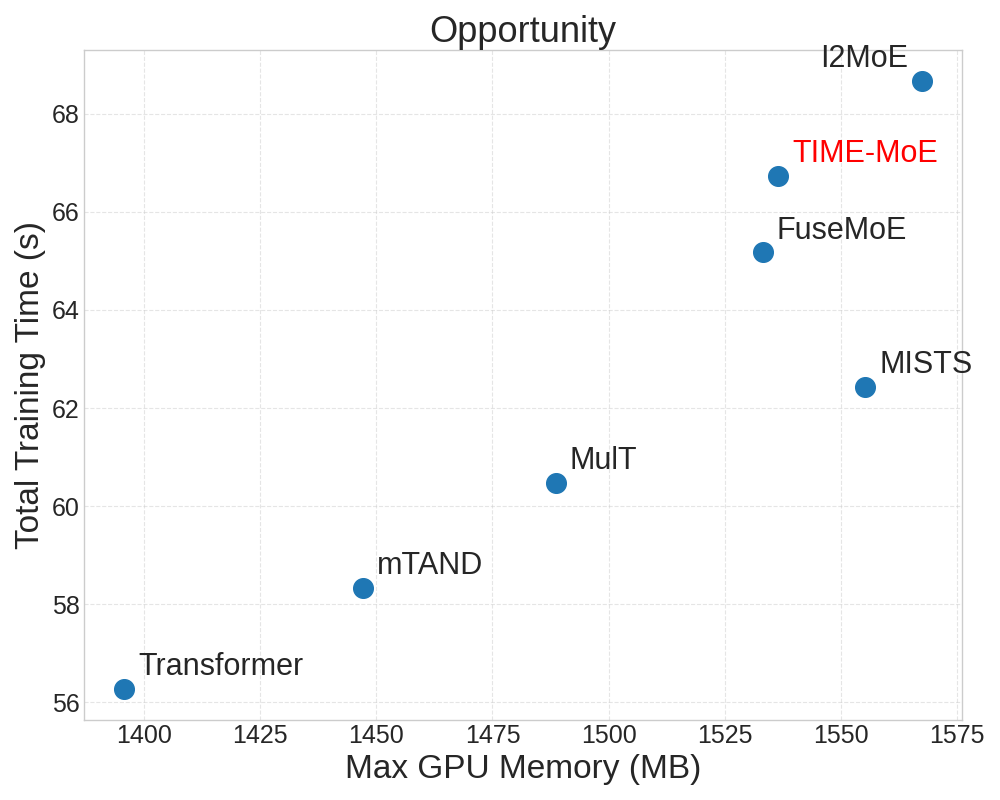}
        %\vspace{-5pt}
        \\
        {\small{(d)}}
        \end{tabular} & 
        \begin{tabular}{c}
        \includegraphics[width=.31\textwidth]{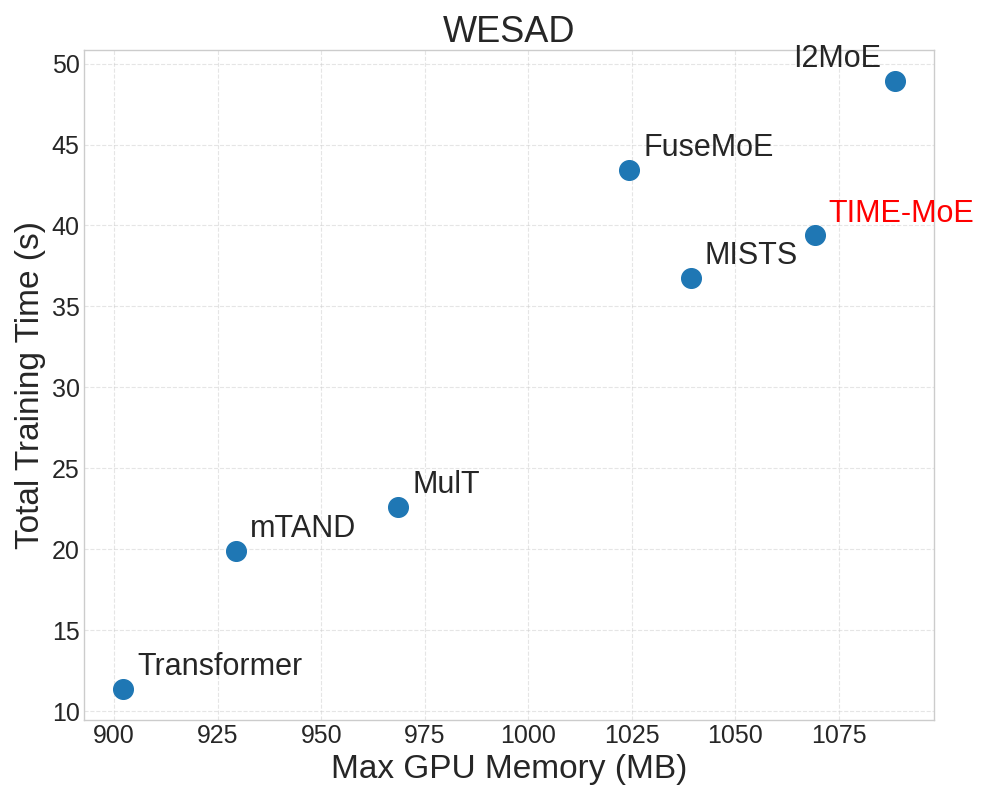} 
        %\vspace{-5pt}
        \\
        {\small{(e)}}
        \end{tabular} &
        \begin{tabular}{c}
        \includegraphics[width=.31\textwidth]{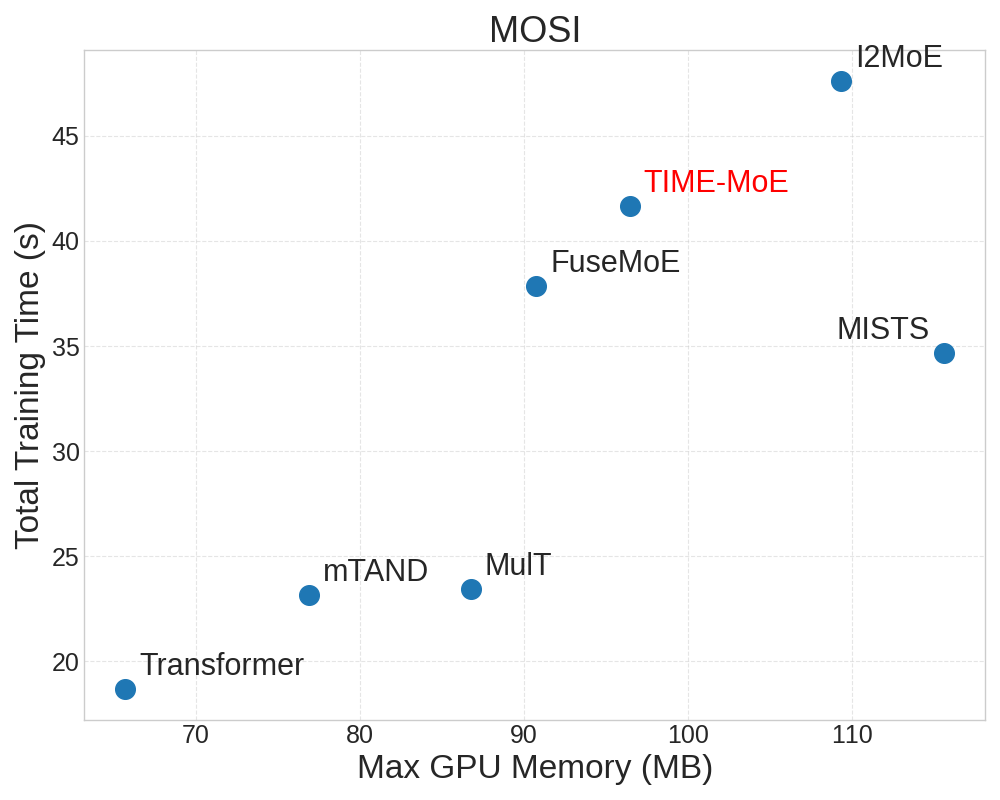} 
        %\vspace{-5pt}
        \\
        {\small{(f)}}
        \end{tabular} \\
        \end{tabular}
    \end{minipage}
    \caption{Comparison of training time and maximum GPU memory usage across all methods and datasets. Although computing and incorporating temporal RUS values into MoE training introduces a certain amount of additional computation and training time, the overall resource requirements of \name~remain comparable to baseline multimodal Transformer/MoE architectures, such as FuseMoE \citep{han2024fusemoe} and MISTS \citep{zhang2023improving}, while being more efficient than I2MoE \citep{xin2025i2moe}. Note that the sequence length of the temporal RUS values varies depending on the task and dataset; longer sequences naturally increase memory consumption. As a result, the relative positioning of \name~in the training-time v.s. memory plot varies across datasets. Models such as the standard multi-head self-attention Transformer, mTAND, and MulT represent earlier-generation attention-based architectures with lower computational requirements, and thus appear on the lower end of the resource spectrum.}
    \label{fig:computation}
\end{figure}

We summarize the total training cost, including both temporal RUS estimation and \name~training, in terms of wall-clock training time and maximum GPU memory consumption across all datasets evaluated in the paper. Figure \ref{fig:computation} presents scatter plots showing the distribution of training time versus peak GPU memory for each dataset.

\subsection{Effect of Class Imbalance in RUS Estimation}
For datasets with sequence-level labels such as MIMIC-IV and MOSI, the prediction task is performed at the full-sequence level. For example, in MIMIC-IV IHM, the model uses the first 48 hours of ICU measurements to predict in-hospital mortality. Accordingly, we compute temporal RUS at the final time step (i.e., sweeping lags backward from the end of the 48-hour window), since these values are the relevant values aligned with the prediction target. However, in situations such as class imbalance and label noise can influence both temporal RUS estimation and downstream routing behavior. As discussed in Section \ref{sec:exp}, each patient contains multiple measurements $\mathcal{D}_i \in \mathbb{R}^{d \times T}$, but is associated with only a single label $Y_i \in \{0, 1, \dots, N\}$. Thus, meaningful estimation of the joint distribution $(X_1, X_2, Y)$ requires a batch of patient trajectories with sufficient label diversity; otherwise the joint distribution becomes degenerate.

\begin{figure}[h] 
\centering 
\begin{tabular}{cc}
%\centering
\begin{tabular}{c}
\includegraphics[width=0.4\linewidth]{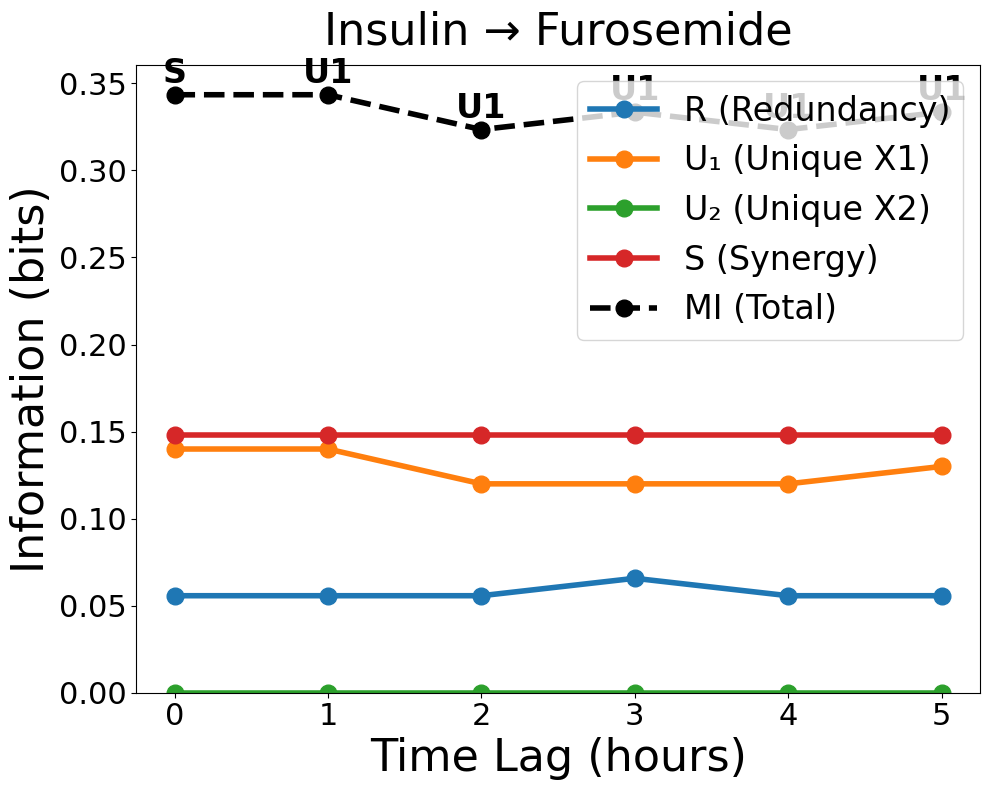}
\\
\hspace{20pt}
{\small{Highly imbalanced batch}}
\end{tabular}
& 
\begin{tabular}{c}
\includegraphics[width=0.4\linewidth]{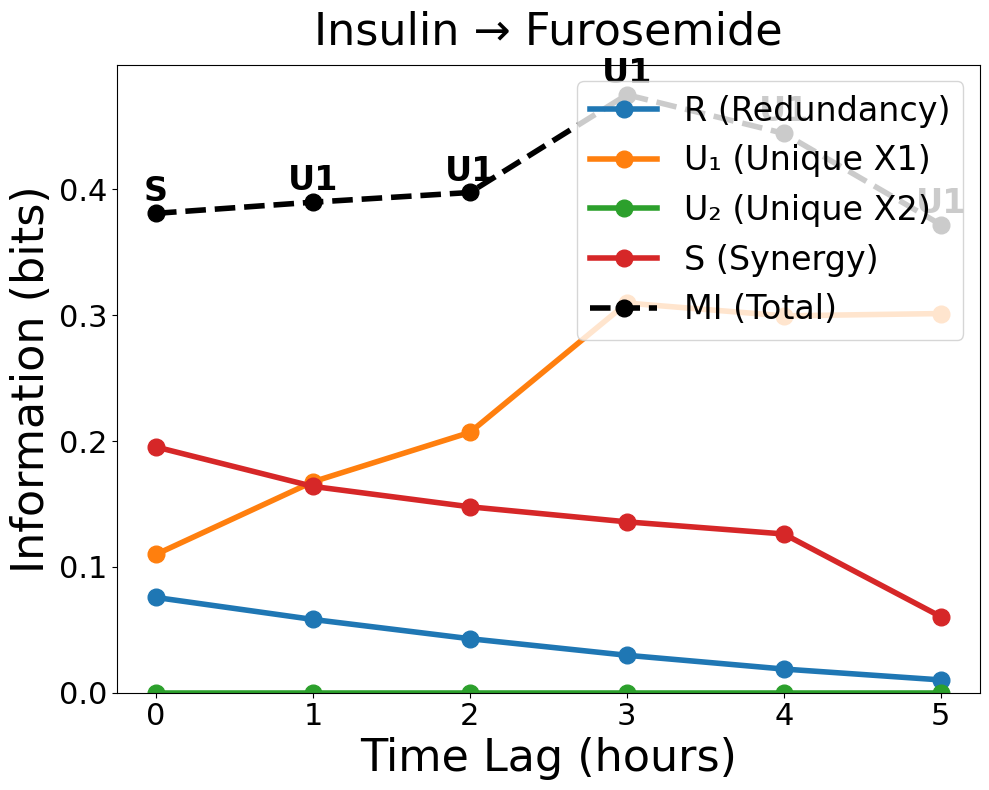}
\\
\hspace{20pt}
{\small{Balanced batch}}
\end{tabular}
\end{tabular}
% \raisebox{5pt}{\rotatebox{0}{\small{Feature Value}}}
%\raisebox{30pt}{\rotatebox{90}{\small{Output}}}
%\subfigure[ Slicing Plot of Monotonic Violation]{\label{fig:slice}\includegraphics[width=.3\textwidth]{figs/slice.pdf}}
\caption{We use the ``Insulin → Furosemide'' modality pair as an example to compare temporal RUS estimations obtained under class-imbalanced settings versus those computed under the regular (balanced) setting. We fed RUS sequences from both \textit{highly imbalanced batch} and \textit{balanced batch} into \name~training. We observed a clear performance degradation for the imbalanced batch: AUROC decreased from 85.40 to 82.35 and F1 score decreased from 84.97 to 81.73.}
\label{fig:imbalance}
\end{figure}

We use MIMIC-IV IHM (in-hospital mortality) as a concrete example to study the importance of class balance and label quality for meaningful temporal RUS estimation. In-hospital mortality is a binary classification task where the label indicates whether the patient dies after ICU discharge, we designed three conditions to evaluate how class imbalance and label quality affect both RUS estimation and model performance:
\begin{enumerate}
    \item A batch containing only patients with outcome 0 or only outcome 1. In this case, temporal RUS cannot be meaningfully computed under the information decomposition framework because $(X_1, X_2, Y)$ becomes degenerate. 
    \item A batch containing both labels but with $>95\%$ belonging to class 0. RUS can be computed, but the resulting temporal RUS sequence is substantially less informative, with the modality interaction over time remaining almost constant. 
    \item A batch with approximately balanced labels and sufficiently rich histories. This corresponds to the original temporal RUS used in the main experiments. 
\end{enumerate}
In Figure \ref{fig:imbalance}, we use the temporal RUS values of the ``Insulin → Furosemide'' modality pair as an illustrative example. Such results provide us insights on the importance of class balance and label quality for meaningful temporal RUS estimation, and demonstrate how degraded RUS estimation can affect the final model performance.
% As shown in Experiment #6 in the link above, we use the temporal RUS values of the “Insulin → Furosemide’’ modality pair as an illustrative example. We then fed these RUS sequences from conditions (2) and (3) into \name~ training. We observed a clear performance degradation for condition (2): AUROC decreased from 85.40 → 82.35 and F1 score decreased from 84.97 → 81.73.

\end{document}